\documentclass[runningheads]{llncs}

\usepackage[T1]{fontenc}
\usepackage{orcidlink}
\usepackage[utf8]{inputenc}

\usepackage[top=80pt,bottom=80pt,left=88pt,right=86pt]{geometry}
\usepackage{graphicx}
\usepackage{array}
\usepackage{algorithm2e}
\usepackage[noend]{algorithmic}
\usepackage{microtype}
\usepackage{subfig}
\usepackage{tabularx}

\usepackage{booktabs}
\usepackage{amsmath}

\DeclareMathOperator*{\argmax}{arg\,max}

\usepackage{enumerate}
\usepackage{tikz}
\usetikzlibrary{shapes.geometric}
\usepackage{wrapfig}
\usepackage{multirow}
\usepackage{nicefrac}
\usepackage[colorinlistoftodos,prependcaption,textsize=tiny]{todonotes}
\usepackage[numbers,square,sort]{natbib}

\usepackage{url}

\renewcommand{\arraystretch}{1.5}

\usepackage{amssymb}
\usepackage{arydshln}
\RestyleAlgo{ruled}
\SetKwComment{Comment}{\# }{}
\SetKwInput{KwData}{Input}
\definecolor{ForestGreen}{RGB}{34,139,34}
\newcommand{\originalarraystretch}{\arraystretch}
\usepackage{sidecap}
\usepackage{caption}

\begin{document}

\title{Patch of Invisibility: Naturalistic Black-Box Adversarial Attacks on Object Detectors}
\titlerunning{Patch of Invisibility}

\author{Raz Lapid\inst{1,2}\orcidlink{0000-0002-4818-9338} \and
Eylon Mizrahi\inst{1} \and
Moshe Sipper\inst{1}\orcidlink{0000-0003-1811-472X}}

\authorrunning{R.~Lapid et al.}

\institute{Dept. of Computer Science, Ben-Gurion University of the Negev,  
Beer-Sheva, 8410501, Israel \\ 
\and
DeepKeep, Tel-Aviv, Israel}

\maketitle
% TODO: Check wherever we talk about ALL the attacks we compare to.
\begin{abstract}
Adversarial attacks on deep learning models have received increased attention in recent years. Work in this area has mostly focused on gradient-based techniques, so-called ``white-box'' attacks, where the attacker has access to the targeted model's internal parameters; such an assumption is usually untenable in the real world. Additionally, some attacks use the entire pixel space to fool a given model, which is neither practical nor physical. To accommodate these problems we propose the \textbf{BBNP} algorithm (Black-Box Naturalistic Patch): a direct, black-box, naturalistic, gradient-free method that uses the learned image manifold of a pretrained, generative adversarial network (GAN) to generate naturalistic adversarial patches for object detectors. This method performs model-agnostic black-box naturalistic attacks on object detection models by relying solely on the outputs of the model. Comparing our approach against five models, five black-box and two white-box attacks, we show that our proposed method achieves state-of-the-art results, outperforming \textbf{all} other tested black-box approaches.
\end{abstract}

\section{Introduction}
\label{sec:intro}

Deep Neural Networks (DNNs) are increasingly deployed in safety-critical applications, many involving some form of identifying humans. The risk of facing deceptive images---so called \textit{adversarial} instances---grows with the use of DNN models. In order to trick the DNN into categorizing the input image differently than a human would, an adversarial sample is employed. As first demonstrated by \citep{szegedy2013intriguing}, neural networks are susceptible to such adversity.

The methods used to generate adversarial instances based on minimal input perturbations were enhanced by subsequent works \citep{carlini2017towards,moosavi2016deepfool,lapid2022evolutionary,lapid2023see}. Instead of modifying existing images, \citep{thys2019fooling} generated unconstrained adversarial instances using generative adversarial networks (GANs). While numerous approaches focus on digital, white-box attacks, wherein the attacker possesses complete access to the model, it is imperative to examine and understand in greater detail the potential realistic attacks that transpire in the real world, characterized by limited or no access to the model by the attacker. Adversarial attacks are a type of cyber-security threat that aims to deceive deep learning (DL) systems by injecting carefully crafted inputs that are designed to fool them. 

These attacks exploit vulnerabilities in DL models and take advantage of their tendency to make mistakes when processing data. Adversarial attacks can be used for manipulation in a wide range of applications, including image recognition, natural language processing, autonomous vehicles, and medical diagnosis \citep{lapid2022evolutionary,vitracktamam2023foiling,lapid2024open}.

The implications of adversarial attacks can be far-reaching, as they can compromise the security and accuracy of systems that rely on DL. For instance, an adversarial attack on a vehicle-mounted, object-detection system could cause it to misidentify a stop sign as a speed-limit sign \citep{eykholt2018robust}, potentially causing the vehicle to crash. As DL becomes increasingly ubiquitous, the need to mitigate adversarial attacks becomes more pressing. Therefore, research into adversarial attacks and defenses is a rapidly growing area, with researchers working on developing robust and secure  models that are less susceptible to such attacks. 

Our research primarily centers on real-world, naturalistic, universal, black-box attacks on object detectors. A naturalistic adversarial patch is a perturbation designed to blend seamlessly with real-world environments, making it difficult to distinguish from benign elements in a scene. These patches introduce targeted misclassifications by mimicking natural variations in texture, color, and lighting, thus enhancing their realism and effectiveness against object detectors. In a previous study, a gradient-based optimization technique was introduced to create effective naturalistic patches that deceive object detectors \citep{hu2021naturalistic}. While that work showcased the adversarial robustness of object detectors, limited consideration was given to the practical scenario where the attacker lacks access to the specific model being targeted. 

The distinctiveness of our approach lies in the utilization of naturalistic patches to adversarially manipulate object detection systems, presenting a notable departure from conventional adversarial noise-based methodologies. This departure is underscored by the inherent alignment of naturalistic patches with the underlying data distribution of the object detector, resulting in a heightened level of stealth. Unlike their adversarial noise counterparts, as seen in \autoref{tab:naturalness}, naturalistic patches seamlessly integrate into the visual context, rendering them less conspicuous and consequently more efficacious in evading detection mechanisms. 

Our work presents a novel contribution in the domain of adversarial AI by employing GANs to produce patches that adversarially perturb object detectors in a black-box setting, circumventing the necessity for gradient information. This approach not only mitigates computational expenses but also reflects a more-realistic scenario wherein attackers lack direct access to the target model. By exploring the latent space of GANs, our methodology offers a unique perspective that enhances the efficacy and applicability of adversarial techniques within real-world contexts, thereby augmenting the current discourse in adversarial AI research.

The main contributions of our work include:

\begin{enumerate}
    \item We introduce \textbf{BBNP} (Black-Box Naturalistic Patch): a novel, efficient, black-box adversarial attack on object-detection frameworks, which assumes a more realistic scenario than did previous works that were all white-box or transfer-based. Those scenarios are far-more demanding to deploy and significantly more computationally expensive.
    
    \item We assess our attack digitally, on five different models, showing that our method outperforms \textbf{all} other black-box methods on all the models tested. 
    
    \item We conduct a comprehensive qualitative assessment of the naturalness of the generated adversarial patches. This includes subjective evaluations through surveys, providing empirical evidence of the realism and effectiveness of our patches in practical scenarios.
\end{enumerate}

\section{Previous work}
\label{sec:prev}
This section offers some background and discusses relevant literature on adversarial attacks. We begin with digital adversarial attacks on object detectors, followed by black-box digital attacks, and ending with GAN-related attacks.

In general, attacks can be divided into three categories: white-box, black-box, and gray-box.
\begin{itemize}%[noitemsep,topsep=1pt]
    \item \textbf{White-box} threat models assume that the attacker has complete knowledge of the system being attacked. This includes knowledge of the system's architecture, implementation details, and even access to the source code. In a white-box threat model, the attacker can inspect the system's internals and use this knowledge to launch targeted attacks.
    
    \item \textbf{Black-box} threat models assume that the attacker has no prior knowledge or access to the system being attacked---no knowledge of the system's architecture or implementation details. This means that the attacker is only able to observe the system's behavior from the outside, without any ability to inspect the internals of the system.
    
    \item \textbf{Gray-box} threat models are a mix of both black-box and white-box models, where the attacker has some knowledge of the system being attacked---but not complete knowledge.
\end{itemize}

Herein we focus on \textbf{black-box} threat models because we only assume access to the models' inputs and outputs.

Convolutional Neural Networks (CNNs) have been of particular focus in the domain of adversarial attacks, due to their popularity and success in imaging tasks \citep{kurakin2018adversarial,feng2021meta,zolfi2021adversarial,lapid2022evolutionary,sharif2019general}. Herein, we shall also focus on CNNs.

\textbf{Digital adversarial attacks on object-detection models}, which are used in a variety of systems, from surveillance devices to autonomous cars, have become an increasing concern. These attacks involve intentionally malicious inputs to fool the model into generating bad predictions, and they can have severe effects, including inaccurate object identification or failing to detect. 

One of the most commonly used object-detection models is You Only Look Once (YOLO) \citep{redmon2016you}, which is based on a single convolutional neural network (CNN) that simultaneously predicts the class and location of objects in an image. Several studies have shown that YOLO is vulnerable to adversarial attacks \citep{liu2018dpatch,im2022adversarial,thys2019fooling,hu2021naturalistic}. For example, targeted perturbation attacks can be used to modify an input image in a way that causes YOLO to misidentify or fail to detect certain objects. An untargeted attack, on the other hand, seeks to create adversarial examples that cause general disruption to the model's performance. Adversarial attacks can have a significant influence on object-detection models. Most prior research regarding such attacks on object-detection models focus on white-box, gradient-based attacks, which is, by and large, a non-real-world scenario. 

\textbf{In the context of black-box digital attacks}, much attention has been given to transfer-based attacks \citep{liu2016delving,li2020towards,zhang2022investigating,qin2021training}. In general, these train a surrogate model, $g_{\theta'}(x)$, which is supposed to approximate the target model, $f_\theta(x)$, i.e., $g_{\theta'}(x) \approx f_\theta(x)$. There are a number of problems with this approach: 
(1) In order to make a suitable approximation we need to know beforehand the training data distribution that the target model was trained on;
(2) we need to know the actual architecture of the target model;
and (3) we must train or finetune the surrogate model, which is computationally expensive. 

Our approach \textbf{does not require knowledge of the actual architecture, nor training of a surrogate model. Thus, it is notably easier and more realistic}. 

Our approach harnesses a pre-trained GAN to craft adversarial patches capable of subverting object detectors in a black-box setting. Unlike surrogate models, which heavily rely on the data distribution used during the training phase of the targeted object detectors, GANs operate within a latent space, dissociated from the specifics of the detector's data distribution. This affords our method the ability to generate patches exhibiting a heightened semblance of naturalness, thereby augmenting their potential to evade detection. The naturalistic nature of the generated patches enhances their stealth, making them potentially indistinguishable when integrated into everyday objects like shirts or signs. Moreover, the optimization of adversarial patches within the latent space of a GAN entails reduced computational demands compared to equivalent operations conducted within pixel space, underscoring a significant advantage of our proposed approach. 

% \textbf{Physical adversarial attacks} pose an even greater threat than digital ones. Most digital adversarial attacks need access to the actual model in order to fool it. Further, many attacks use global perturbation over pixel space---e.g., changing the sky's pixels---thus rendering it less realistic. Physical adversarial attacks can be engendered in a variety of ways, including covering an object with paint or some other material \citep{wang2021dual,brown2017adversarial}, applying stickers \citep{eykholt2018robust,komkov2021advhat} or camouflage \citep{duan2020adversarial,wang2021dual}, or changing the object's shape or texture \citep{hu2022adversarial,yang2020patchattack}. Such changes are intended to alter how the object appears to the model while maintaining it recognizable to humans. 
% Interestingly, the attack often looks identical to the naked eye---but \textit{not} to the model (as we shall also show below).

% Physical adversarial attacks can have serious effects, e.g., when an autonomous car's failed detection results in a crash. Physical attacks can also be intentionally used to get past security systems or enter restricted areas without authorization. Again, most prior research on physical attacks has been done using gradients, in a white-box setting. 

Contrarily, we create \textbf{adversarial patches} in a \textbf{black-box manner}, i.e., without the use of gradients, nor surrogate-models, leveraging the \textbf{learned image manifold of GANs}, making our attack model-agnostic \citep{goodfellow2020generative}.

\textbf{Generative Adversarial Networks (GANs)}:
The quality of generative models used for image generation has improved significantly with the advent of GANs \citep{goodfellow2020generative} and Diffusion Models \citep{ho2020denoising}. Herein, we focus on GANs, due to their relatively small latent-space dimension and their fast image generation. GANs utilize a learnable loss function, where a separate discriminator network is used to distinguish between real and fake images, and the image generator network is trained to synthesize images that are indistinguishable from real ones. Despite their visually appealing results, GANs often face issues such as instability, vanishing gradients, and mode collapse. Researchers have suggested many different GANs to address these issues \citep{karras2019style,arjovsky2017wasserstein,mao2017least}. Herein we chose to use BigGAN2 \citep{brock2018large} due to its relatively small latent dimension $(d=128)$, its class-conditional generation, and its ability to generate high-quality and high-resolution images.

\section{Method}
\label{sec:method}
Our goal is to generate \textit{universal} adversarial patches that are both effective and appear \textit{naturalistic}, while avoiding the use of gradients and surrogate models. An adversarial patch is a specific type of attack where a small, local pattern is added to an image to cause a model to make incorrect predictions.  Formally, given a model $f_\theta$, a dataset $\mathcal{D}$ of \textit{(image, label)} pairs, a loss function $\mathcal{L}$, and a norm $\Vert \cdot \Vert_p$, the objective of a \textit{universal adversarial attack} is formulated as follows:
\begin{equation}
    \argmax_{\Vert \delta \Vert_p \leq \epsilon} \mathbb{E}_{(x, y)\in \mathcal{D}} \mathcal{L}\left[f_\theta (x + \delta), y\right],
\end{equation}
where $\delta$ is the perturbation added to the clean input images $(x,y) \in \mathcal{D}$, and $\epsilon$ is the maximal allowed perturbation. Essentially, we aim to find a perturbation, $\delta$, with norm not exceeding $\epsilon$, such that it maximizes the loss $\mathcal{L}$.
When added to the clean images, the model $f_\theta$ is eventually induced to make a mistake. Our algorithm uses the above objective, as seen in \autoref{es-algorithm}.

The challenge lies in the vast space of possible adversarial patches. To make the search process more efficient, we utilize pretrained GANs. By searching through the latent space of a GAN, we can craft adversarial perturbations more effectively. The latent space of the GAN is significantly smaller than the pixel space of images, which simplifies the optimization process. Additionally, the images generated by the GAN often look more natural compared to directly optimizing in pixel space, making such attacks harder to detect \citep{brown2017adversarial,lee2019physical}.

Given a pretrained GAN generator \( g \) and an object detector \( f \), our objective is to find a latent vector \(\textbf{z}\) that, when used as input to \( g \), produces an adversarial patch \( g(\textbf{z}) \). This patch, when applied to an image \( x \), should deceive the object detector. Specifically, we seek:

\begin{itemize}
    \item An image \( x_{\text{adv}} = x \bigoplus g(\textbf{z}) \) where the patch is applied to specific regions of \( x \) (\(\bigoplus\) denotes this patch application).
    \item A significant alteration in the object detector's output, so that \( f(x_{\text{adv}}) \neq y \), where \( y \) is the true label for \( x \).
\end{itemize}

We exploit the smaller dimensionality of the GAN's latent space by using an Evolution Strategy algorithm \citep{wierstra2014natural}. This method updates the latent vector \(\textbf{z}\) iteratively based on feedback from the fitness function. To compare, optimizing a patch directly in pixel space (of size \(3 \times 256 \times 256\), as in our experiments) would involve \(196,608\) pixels. In contrast, our approach operates in a \(128\)-dimensional latent space, achieving a reduction in dimensionality by a factor of \(1{,}000\). This reduction makes the optimization process faster and the generated attacks more natural, thanks to the pretrained GAN, which is trained on ImageNet \citep{deng2009imagenet}.

\begin{figure}[ht]
\centering
\includegraphics[trim={3cm 3cm 3cm 3cm},clip,scale=0.5]{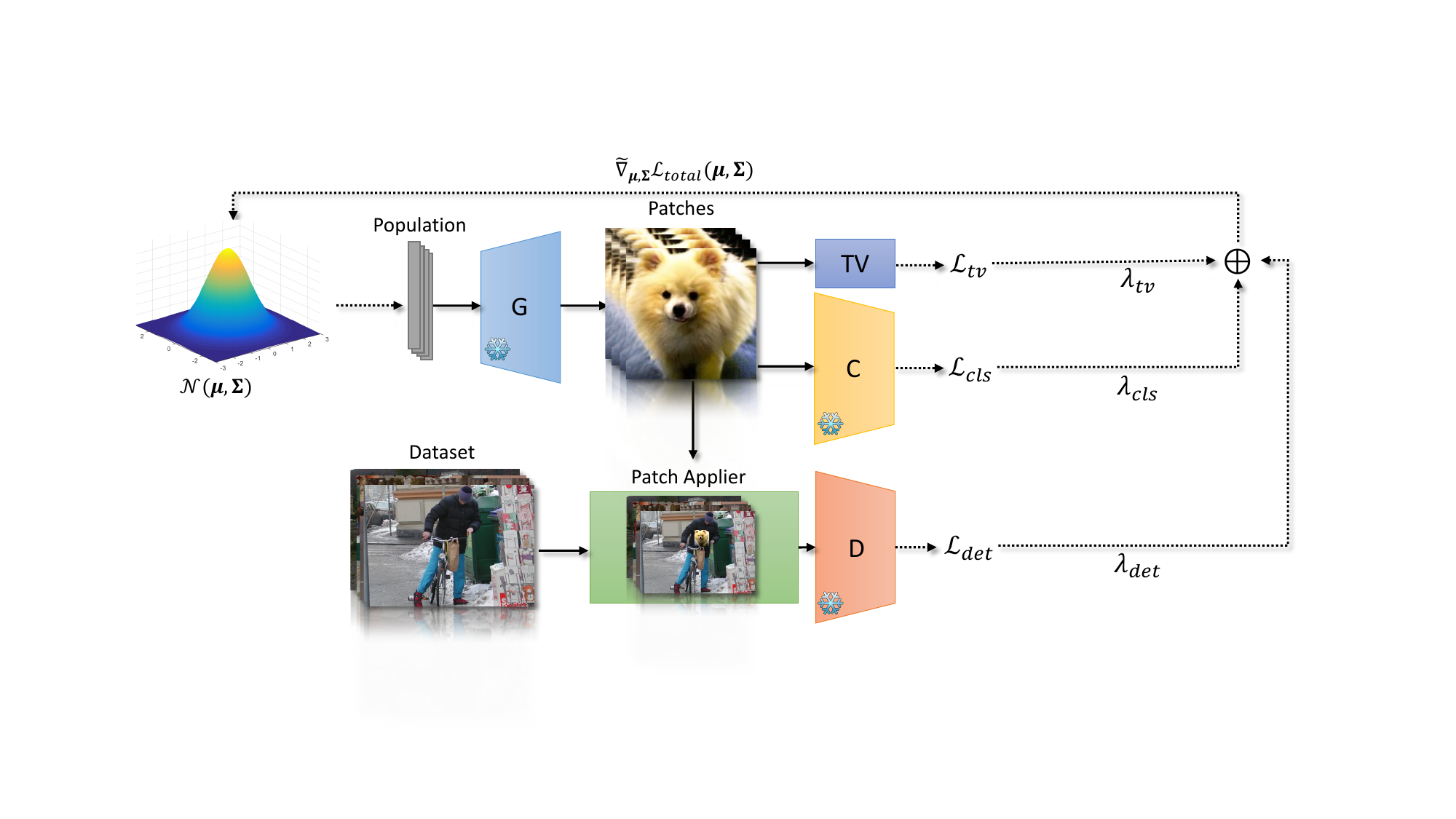}
\vspace{-20pt}
\caption{Naturalistic Black-Box Adversarial Attack: Overview of framework. The system creates patches for object detectors by using the learned image manifold of a pretrained GAN generator ($G$) on real-world images. We use a pretrained classifier ($C$) to force the optimizer to find a patch that resembles a specific class, the $TV$ component, which refers to total variation, aims to ensure the images are rendered with maximal smoothness, and the detector ($D$) for the actual detection loss. Efficient sampling of the GAN images via an iterative evolution strategy ultimately generates the final patch. The snowflake icon means frozen model weights.}
\label{fig:algorithm}
\end{figure}

\autoref{fig:algorithm} illustrates the overall approach. We aim to find a latent vector that produces a patch capable of causing the object detector to fail to recognize targets consistently, making it a universal adversarial patch.

% We leverage the latent space's (relatively) small dimension, approximating the gradients using an Evolution Strategy algorithm \citep{wierstra2014natural}, repeatedly updating the input latent vector by querying the target object detector until an appropriate adversarial patch is discovered. For comparison, if we were to optimize a patch directly in pixel space of size $3 \times 256 \times 256$, which is the patch size we've optimized using GANs, we'd need to optimize $3 \times 256 \times 256 = 196608$ pixels instead of $128$ features in the GAN's latent space. Working in latent space achieves a $1000\times$ dimensionality reduction of the space of optimization. Further, the attacks look more natural because we sample from a pretrained GAN that is trained on ImageNet \citep{deng2009imagenet}.

% \autoref{fig:algorithm} depicts a general view of our approach. We seek an input latent vector that, when processed by a pretrained generator, produces an image that makes the object detector incorrectly fail to recognize any persons with this patch, thus making it a universal adversarial patch.

% \newpage

\autoref{es-algorithm} presents the pseudocode for the Evolution Strategy. In each iteration \( t \), we generate \( n \) Gaussian noise vectors \(\{\boldsymbol{\epsilon_1}, \boldsymbol{\epsilon_2}, \ldots, \boldsymbol{\epsilon_n}\}\) and add them to the current latent vector \(\boldsymbol{z_t}\). We then feed these perturbed latent vectors into the generator \( g \) and update \(\boldsymbol{z_{t+1}}\) based on the resulting losses. This process is straightforward but highly effective in creating adversarial patches for object detectors.

% \autoref{es-algorithm} shows the pseudocode of the evolution strategy.
% In each iteration $t$, we sample $n$ Gaussian noise vectors, $\boldsymbol{\epsilon_{1}, \epsilon_{2},\ldots,\epsilon_{n}}$ (there is one latent vector per noise value). Then, we scale them using $\sigma$, add them to the latent vector $\boldsymbol{z_{t}}$, and feed to the pretrained generator $g$. The latent vector $\boldsymbol{z_{t+1}}$ is then updated using the weighted sum of the loss values $\mathcal{L}_i$ for each $\boldsymbol{\epsilon_i}, i\in \{1,2,...n\}$. Although the attack generation process outlined in \autoref{es-algorithm} is straightforward, we will demonstrate its effectiveness in attacking object detectors.

\begin{algorithm}
\scriptsize
\caption{BBNP}\label{alg:two}
\KwData{pretrained generator $g$, object detector $f$, loss function $\mathcal{L}$, step size $\alpha$, noise standard deviation $\sigma$, initial latent vector $\boldsymbol{z_0}$, number of iterations $T$, population size $n$, dataset $\mathcal{D}$}
\For{$t = 1,\ldots,T$}{
    Sample $\boldsymbol{\epsilon_{1}}, \boldsymbol{\epsilon_{2}},\ldots,\boldsymbol{\epsilon_{n}}\sim \mathcal{N}(\textbf{0}, \textbf{I})$ \\
    \For{$i = 1, \ldots, n$} { 
        $\mathcal{L}_i = \mathbb{E}_{(x, y)\in \mathcal{D}} \left[ \mathcal{L}(f(x \bigoplus g(\boldsymbol{z_t} + \sigma \boldsymbol{\epsilon_i})), y) \right]$ \\
    }
    Set $z_{t+1} \leftarrow \alpha \frac{1}{n \sigma} \sum^{n}_{i=1}\mathcal{L}_{i} \boldsymbol{\epsilon_i}$
}
\label{es-algorithm}
\end{algorithm}

\subsection{Generating adversarial patches}
Previous research optimized adversarial patches in \textit{pixel space}. 
We, on the other hand, focus on a GAN generator's \textit{latent space}. Our resultant adversarial patch will be closer to the manifold of natural pictures and hence appear more realistic (because GANs learn a latent space that roughly approximates the manifold of natural images).
We employ a generator $G$ that has been previously trained on ImageNet using a GAN framework, and we search the space of learned natural image manifold using an evolution strategy.

The evolution strategy algorithm begins by using an isotropic standard multivariate distribution, $\mathcal{N}(\boldsymbol{\mu}=\textbf{0}, \boldsymbol{\Sigma}=\textbf{I}_d)$, which parameterizes the evolved population with $\boldsymbol{\mu}$ and $\boldsymbol{\Sigma}$. We begin with an initial latent vector $\boldsymbol{z_0}$. We then randomly sample $n$ noises $\boldsymbol{\epsilon_1, \epsilon_2,...,\epsilon_n}\in \mathbb{R}^d$ using the standard multivariate distribution, scale them by $\sigma$ and add them to the initial latent vector $\boldsymbol{z_0}$ resulting with $\textbf{Z}\in \mathbb{R}^{d\times n}$ --- which then are fed to the generator to create the population of patches $P=g(\textbf{Z})\in \mathbb{R}^{n \times 3 \times h \times w}$ -- where $n$ stands for the population size, 3 is the number of channels (RGB), $h$ is the patch's height, and $w$ is the patch's width.

Then, using gradient approximation through an evolution strategy, we repeatedly search the latent vector $z$ that best achieves our objective function, which is:

\begin{equation}
    \mathcal{L}_{total} = \mathcal{L}_{det} + \lambda_{tv} \mathcal{L}_{tv} + \lambda_{cls} \mathcal{L}_{cls}, 
\end{equation}

where:

\begin{itemize}%[noitemsep,topsep=1pt]
    \item $\mathcal{L}_{det}$: detection loss of object detector of the specific class.
    \item $\mathcal{L}_{tv}$: total variation loss, to promote smoothness of generated patch. The loss function is as follows:
    \begin{equation}
        \mathcal{L}_{tv} = \sum_{i,j} \left( \left| p_{i,j} - p_{i+1,j} \right| + \left| p_{i,j} - p_{i,j+1} \right| \right),   
    \end{equation}
    to promote the generation of visually coherent adversarial patches. The TV loss penalizes abrupt changes in pixel intensity, encouraging smoother and more realistic patches.
    \item $\mathcal{L}_{cls}$: classification loss, to promote more-realistic patch generation.
    \item $\lambda_{tv}$ and $\lambda_{cls}$ are regularization weights.
\end{itemize}

We expound upon these terms below.

\subsection{Adversarial gradient estimation} \label{loss_subsection}
In order to generate patches that may deceive the target object detector, the generator uses adversarial gradient estimations as its primary navigation tool. We initially add the patch onto a given image. In order to compute an adversarial loss for the detection of bounding boxes (BBs), we feed the image to the object detector.

\begin{figure}[ht]
\centering

\begin{tabular}{ccc}
    \includegraphics[width=0.3\textwidth]{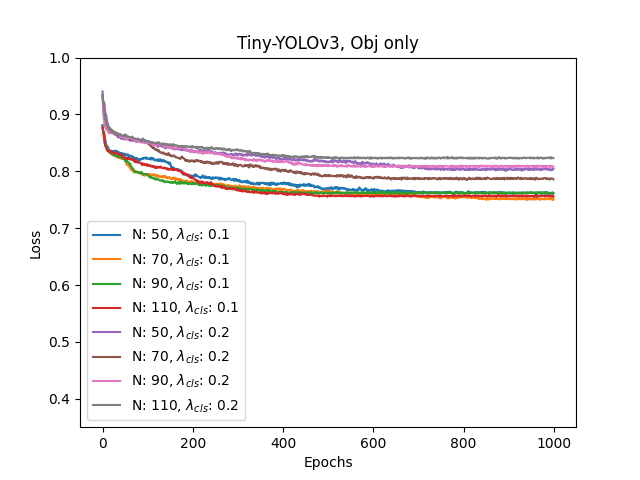} &
    \includegraphics[width=0.3\textwidth]{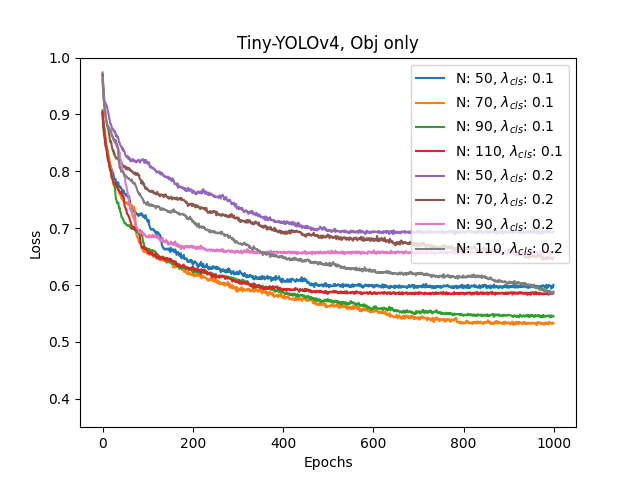} &
    \includegraphics[width=0.3\textwidth]{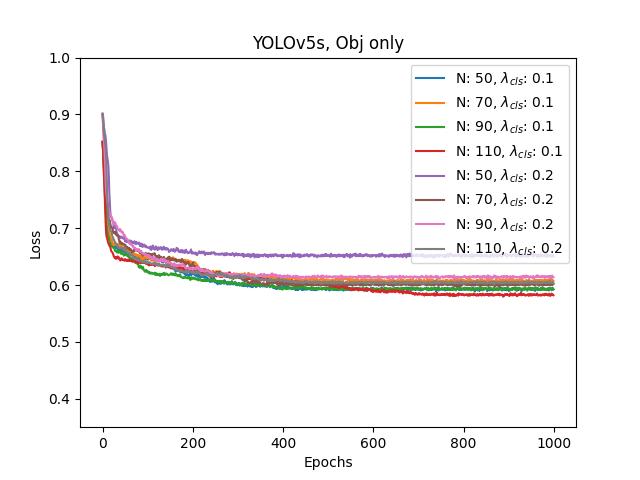} \\
    \includegraphics[width=0.3\textwidth]{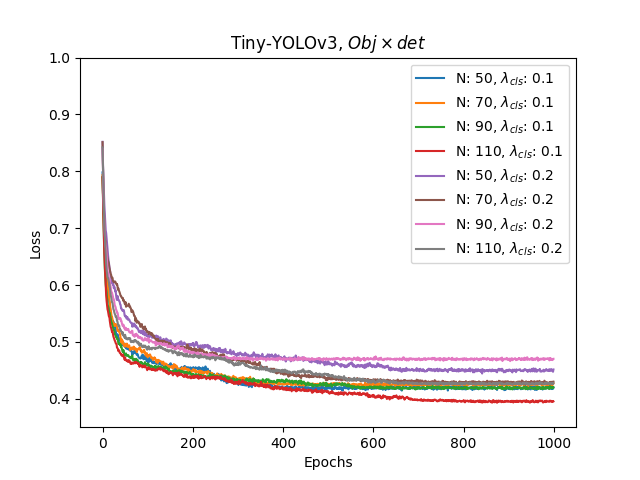} &
    \includegraphics[width=0.3\textwidth]{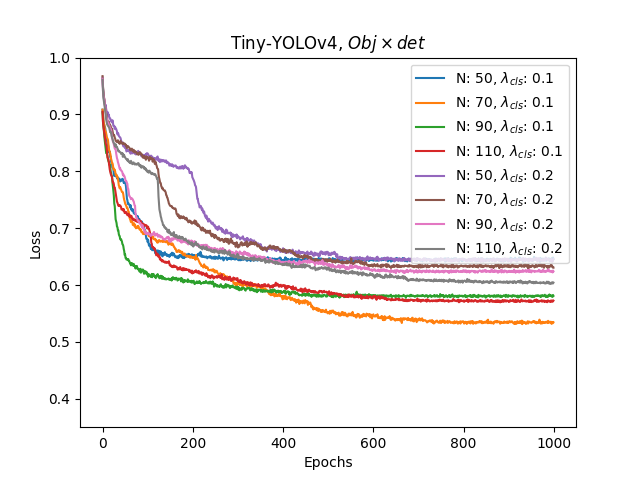} &
    \includegraphics[width=0.3\textwidth]{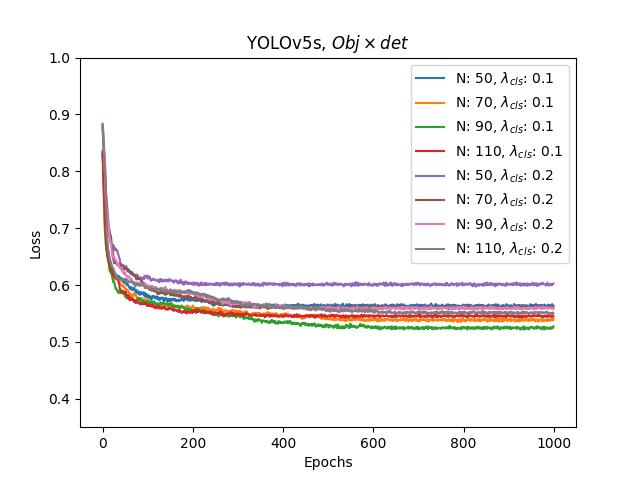} \\
\end{tabular}
\vspace{-10pt}
\caption{Comparison of the effectiveness of the objectness-only loss function versus the objectness $\times$ classification ($obj \times det$) loss function. The first row illustrates the results using the $obj$-only loss function, while the second row demonstrates the performance with the $obj \times det$ loss function. Empirical results consistently show that the $obj \times det$ approach outperforms the objectness-only method.}
\label{fig:losses-comparison}
\end{figure}

% Need to rewrite this paragraph
% skipping, MS
\textbf{Adversarial detection loss}. Detection can be arbitrarily produced by object detectors like YOLO \citep{redmon2016you}. We are interested in minimizing two terms for the patch detection $i$: its objectness probability $D^{i}_{obj}$, which specifies the model's confidence regarding whether there is an object or not, and the class probability $D^{i}_{cls}$, which specifies the model's confidence of a specific class. In this paper we focused on generating patches that conceal humans. Thus, we want to minimize both the objectness $D^{i}_{obj}$ and class probabilities $D^{i}_{cls}$ for our generated patch with respect to the person class, i.e., minimizing the term:

\begin{equation}
\label{det_loss_eq}
    \mathcal{L}_{det} = \frac{1}{n}\sum_{i=1}^n \sum_{j=1}^{|det|} D^j_{obj}(x_i)D^j_{cls}(x_i),
\end{equation}
where $n$ is the population size and $|det|$ is the number of human detections by the model.
By minimizing $\mathcal{L}_{det}$, we achieve a patch that minimizes the objectness and class probabilities of the target class.  During our experiments, we also explored minimizing the objectness probability $D^j_{obj}$ alone. However, we observed that this approach yielded inferior results compared to minimizing the product of the objectness and class probabilities. The addition of class probabilities better captures the model's overall confidence in detecting the target class, thereby enhancing the effectiveness of the generated patches. This observation is substantiated by the plots shown in \autoref{fig:losses-comparison}, which illustrate the performance differences between these approaches.

% \textbf{Physical transformations}. We have no influence over the adversarial patch's viewpoint, position, or size, with respect to the images. In order to enhance the robustness of our adversarial patch, we overlay it onto a human and generate a variety of settings with distinct configurations. Furthermore, we subject our created adversarial patch $P$ to various transformations, including rotation, brightness change, and resizing, to mimic the different visual appearances it may adopt in real-world situations. The transformations are applied by the Patch Transformer (\autoref{fig:algorithm}).  

\textbf{Smoothness}. To promote smoothness in the generated image we apply the term $\mathcal{L}_{tv}$, which represents the total variation loss. 
The calculation of $\mathcal{L}_{tv}$ from a patch $p\in \mathbb{R}^{3 \times h \times w}$ is done as follows:

\begin{equation}
    \mathcal{L}_{tv}(p) = \sum_{i,j} \sqrt{(p_{i+1,j}-p_{i,j})^2+(p_{i,j+1}-p_{i,j})^2},
\end{equation}

where subscripts $i$ and $j$ refer to the pixel coordinates of patch $p$. A constant value of $\lambda_{tv} = 0.1$ was employed in all experiments presented in this paper.

\textbf{Classifier guidance}. Empirically, we observed that the patches generated by the generator often exhibit highly abstract characteristics, diverging significantly from the latent image manifold encoded by the generator. To mitigate this issue and enhance the resemblance of the patch to a specific class, we incorporated an additional loss term. Specifically, we introduced a pretrained ResNet-50 classifier \citep{he2016deep}, trained on the ImageNet dataset \citep{deng2009imagenet}, as a regularization mechanism. This classifier serves to guide the patch towards a more class-specific appearance. It is important to note that the assumption of the attacker having access to a local pretrained classifier, such as ResNet-50, is a standard and widely accepted premise in adversarial attack scenarios, given the ubiquity of such pretrained models in practice.

\begin{figure}[ht]
    \centering
    \begin{tabular}{cccccccc}
        \includegraphics[width=0.1\linewidth]{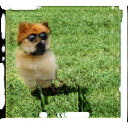} & 
        \includegraphics[width=0.1\linewidth]{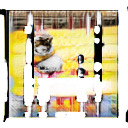} &
        \includegraphics[width=0.1\linewidth]{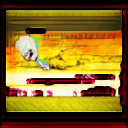} &
        \includegraphics[width=0.1\linewidth]{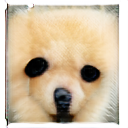} &
        \includegraphics[width=0.1\linewidth]{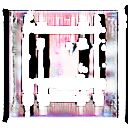} &
        \includegraphics[width=0.1\linewidth]{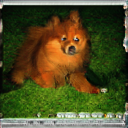} &

        \includegraphics[width=0.1\linewidth]{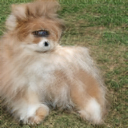} &
        \includegraphics[width=0.1\linewidth]{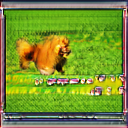} \\
    \end{tabular}
    % Optionally, you can add a caption for the whole table here.
\vspace{-10pt}
    \caption{Images generated utilizing BigGAN, without applying any projection onto the norm of the latent vector.}
    \label{fig:naturalness1}
\end{figure}

\textbf{Naturalness}. To maintain naturalness we enforce a constraint on the norm of the latent vectors $\textbf{Z}$, which should not exceed a threshold $\tau$. By adjusting this threshold we can balance naturalness versus attack performance. We use $\Vert \cdot \Vert_{\infty}$ to constrain $\textbf{z}$, resulting in:

\begin{equation}
    \textbf{z}^t = \pi(z^{t-1} - \alpha \tilde{\nabla}_{\textbf{z}} \mathcal{L}_{total}),
\label{eq:projection}
\end{equation}
where:
\begin{equation}
    \pi (\textbf{z}) = \{z_i | z_i \leftarrow min(max(z_i, -\tau), \tau), \forall z_i \in \textbf{z}\},
\end{equation}
$t$ is the timestep (epoch), $\alpha$ is the step size, and $\tilde{\nabla}_{z} \mathcal{L}_{total}$ is the gradient approximation of the total loss with respect to the latent vector $z$. We used $\tau=10$ in all experiments based on qualitative observations as follows. \autoref{fig:naturalness1} displays the patches generated without any form of projection. Conversely, \autoref{fig:patches} presents the patches generated through the application of \autoref{eq:projection}.

In \autoref{sec:naturalness_assessment}, we employed a subjective evaluation paradigm to assess the perceived naturalness of the generated patches.

Our methodology does not involve searching the latent space for an optimal patch. Instead, we use the most effective patch from the training dataset optimization and evaluate its performance on the test dataset.

\section{Experiments and Results}
\label{sec:results}
We begin by delineating the implementation details of our experiments, followed by the quantitative and qualitative experiments themselves, focusing on the effectiveness of the proposed adversarial patch using the INRIA person dataset \citep{dalal2005histograms}, where we target images with one person, to facilitate optimization; note that while we used single-person images for optimization, testing was done on the \textit{entire} dataset, which includes both single and multi-person images.
We also performed several ablation studies (\autoref{subsec:ablation}) to refine hyperparameters.
We also conducted transferability checks against five different models, five black-box attacks \citep{croce2022sparse,andriushchenko2020square,ilyas2018black,bergstra2012random} and two white-box attacks \citep{hu2021naturalistic,thys2019fooling} to assess the generality of our attack. To further evaluate the naturalness of the patches, we also included a subjective human survey, which can be found in \autoref{sec:naturalness_assessment}. The code will be available on GitHub at \url{anonymized}. % https://github.com/razla/Patch-of-Invisibility

\subsection{Experimental setup}
The experiments were performed using the Adam \citep{kingma2014adam} optimizer, with a learning rate of $0.02$ and $\beta_1=0.5, \beta_2=0.999$. The optimization process involved populating the gradients through our evolutionary algorithm. If the changes in losses remain below $1e-4$ for at least 50 epochs, we reduce the learning rate, following \citep{hu2021naturalistic}. Our generator consists of BigGAN2 \citep{brock2018large} (with frozen weights), which has been pretrained on the ImageNet dataset, with a latent vector of size $120$, and $128 \times 128$ output resolution. As the BigGAN generator is class-conditional, the generated patch can be guaranteed to belong to a specific class. We used five different object detectors---Tiny-YOLOv3 \citep{redmon2018yolov3}, Tiny-YOLOv4 \citep{bochkovskiy2020yolov4}, YOLOv5s \citep{Jocher_YOLOv5_by_Ultralytics_2020} \citep{redmon2016you,terven2304comprehensive}, SSD-L \citep{howard2019searching,sandler2018mobilenetv2} and L-DETR \citep{li2023lite} trained on the COCO dataset \citep{lin2014microsoft}---with an input resolution of $416 \times 416$. We chose these faster, memory-efficient models so as to be able to conduct more experiments. Moreover, these models represent architectures that are predominantly deployed on edge hardware and widely utilized in production environments \citep{feng2022benchmark,ali2021face,gupta2022edge}, thereby enhancing the practical relevance and realism of the experiments conducted. For each of the detected bounding boxes, we placed an adversarial patch of size $25\%$ of the bounding box. We adopt the method proposed by \citet{hu2021naturalistic} and generate a mask encompassing the clothing region for each individual, onto which we position our adversarial patch.

In our case, we evaluated the Average Precision (AP) specifically for the person class. The mean average precision (mAP) is defined as:

\begin{equation}
\text{mAP} = \frac{1}{N} \sum_{k=1}^{N} \text{AP}(k),
\end{equation}

where \( N \) is the number of object classes and \(\text{AP}(k)\) is the average precision for class \( k \). The average precision for a single class is computed as:

\begin{equation}
\text{AP}(k) = \frac{1}{\text{n}} \sum_{i=1}^{\text{n}} P(i)
\end{equation}

where \( \text{n} \) is the number of objects for class \( k \), \( P(i) \) is the precision at the \( i \)-th object, according to a specific Intersection over Union (IoU) threshold. For this calculation, we set the IoU threshold to 0.5.

The AP values of all experiments are shown in \autoref{table:mAPall} (for reference we also show no-attack values).

We compared our proposed attack to five other black-box methods and two white-box methods: Square Attack \citep{andriushchenko2020square}, which exploits the CNN vulnerability to square perturbations; Sparse-RS \citep{croce2022sparse}, which leverages \citep{andriushchenko2020square} to craft minimal modifications of squares;
RS (pixel space random search) \citep{croce2022sparse}, which uses random search on pixel space;  GAN Latent RS (latent-space random search) which uses random search in GAN latent space;
and NES (natural evolution strategies) \citep{ilyas2018black}, which is an evolutionary attack performed in pixel space. We employed a consistent query budget of 110,000 for all attacks, derived from our chosen population size of 110 and 1,000 epochs. \footnote{We slightly adjusted some of the compared methods to accommodate the object-detection task: We removed the limitation on $\epsilon$ to encourage broader exploration during optimization in competitive attacks, because perturbation limits are not part of our targeted scenario, which is patch attacks.} In our implementation of Square Attack, Sparse-RS, and the two white-box attacks, WB-Nat and WB-Adv as proposed by \citet{thys2019fooling} and \citet{hu2021naturalistic}, we adhered strictly to the default parameters specified in their respective works. For NES, RS, and Latent-RS, we employed a noise standard deviation of $\sigma=0.1$. Additionally, for Latent-RS, we utilized two distinct classifier guidance values, specifically 0.1 and 0.2, in alignment with the parameters used in our proposed attack algorithm.

\newcolumntype{C}[1]{>{\centering\arraybackslash}m{#1}}	
\definecolor{Gray}{gray}{0.9}

\begin{SCtable}[\sidecaptionrelwidth][ht] % [\sidecaptionrelwidth] controls the width of the caption
  \centering
    \caption{Different evaluations of patches in terms of AP(\%) for the INRIA dataset using \autoref{det_loss_eq}. AP is evaluated using best evolved patch per experiment. For each black-box experiment, the best AP (lowest) reached is in \textbf{bold}. White-box attacks are greyed out since they adhere to a completely different set of assumptions, as discussed in text.}
    \label{table:mAPall}
  % \begin{minipage}{0.48\textwidth}
    % \resizebox{169px}{!}{
    % \begin{tabular}{m{1.5cm} | m{1.5cm} m{1.5cm} | m{1.5cm} m{1.5cm} m{1.5cm}}%{@{}l|cc|ccc@{}}
    \renewcommand{\arraystretch}{1.1}
    \resizebox{0.7\linewidth}{!}{
    \begin{tabular}{C{1.5cm} | C{1cm} C{1cm} | C{2cm} C{2cm} C{2cm} C{2cm} C{2cm}} 
        % \toprule
        Attack & $\lambda_{cls}$ & Pop \hspace{3px} & Tiny-YOLOv3 \hspace{3px} & Tiny-YOLOv4 \hspace{3px} & YOLOv5s & SSD-L & L-DETR \\
        \midrule
        \multirow{8}*{\textbf{BBNP}} & \multirow{4}*{$0.1$} & $50$ & $26.1$ & $31.4$ & $33.9$ & $39.8$ & $62.8$ \\  
        &  & $70$ & $31.5$ & $ \textbf{16.3}$ & $31.3$ & $37.9$ & $64.0$ \\  
        &  & $90$ & $24.6$ & $16.9$ & $\textbf{28.6}$ & $39.9$ & $57.8$ \\  
        &  & $110$ & $\textbf{23.2}$ & $20.7$ & $33.3$ & $\textbf{37.2}$ & $\textbf{57.3}$ \\  
        \cline{2-8}
        &  \multirow{4}*{$0.2$} & $50$ & $29.7$ & $31.0$ & $34.6$ & $42.2$ & $73.8$ \\  
        &  & $70$ & $28.9$ & $33.3$ & $33.4$ & $43.1$ & $65.7$ \\  
        &  & $90$ & $33.4$ & $32.7$ & $31.5$ & $44.1$ & \textbf{$60.7$} \\  
        &  & $110$ & $30.0$ & $18.8$ & $33.0$ & $39.5$ & $77.7$ \\
        \hline
        Sparse-RS & N/A & N/A & $41.6$ & $47.1$ & $63.0$ & $64.0$ & $82.2$ \\
        \hline
        Square & N/A & N/A & $43.9$ & $61.6$ & $61.7$ & $62.8$ & 86.3 \\
        \hline
        \multirow{4}*{NES} & \multirow{4}*{N/A} & $50$ & $79.8$ & $77.9$ & $76.0$ & $67.7$ & $89.1$ \\ 
        &  & $70$ & $77.7$ & $78.6$ & $75.4$ & $67.7$ & $87.7$ \\
        &  & $90$ & $82.4$ & $78.4$ & $71.8$ & $67.5$ & $88.9$ \\
        &  & $110$ & $79.7$ & $78.2$ & $75.7$ & $67.2$ & $87.7$ \\
        \hline
        RS &  N/A & N/A & $79.6$ & $79.6$ & $73.6$ & $67.4$ & $88.7$ \\
        \hline
        \multirow{2}*{Latent-RS}  &  $0.1$ & N/A & $68.2$ & $67.1$ & $53.5$ & $67.6$ & $87.0$ \\
         &  $0.2$ & N/A & $68.2$ & $67.1$ & $54.0$ & $61.9$ & $87.0$ \\
        \hdashline[2pt/2pt]
         WB-Nat & N/A & N/A & \textcolor{gray!90}{$16.9$} & \textcolor{gray!90}{$14.5$} & \textcolor{gray!90}{$36.1$} & \textcolor{gray!90}{$35.9$} & \textcolor{gray!90}{86.6} \\
         WB-Adv & N/A & N/A & \textcolor{gray!90}{$8.4$} & \textcolor{gray!90}{$11.3$} & \textcolor{gray!90}{$25.6$} & \textcolor{gray!90}{$43.7$} & \textcolor{gray!90}{85.9} \\
        \hdashline[2pt/2pt]
        % \hdashline
         No Attack &  N/A & N/A & $100.0$ & $93.0$ & $94.0$ & $72.4$ & $95.4$ \\
        % \bottomrule
        \end{tabular}}

        \renewcommand{\arraystretch}{\originalarraystretch} % Reset to saved value
    % }    
  % \end{minipage}
  % \hfill
  % \begin{minipage}{0.45\textwidth}
  %   \centering
  %   \resizebox{175px}{!}{

  %       }
  %   \caption{Transferability results of the different adversarial patches across the different object detection models in terms of mAP (\%). Each row corresponds to a different source model, while the columns represent the target model.}
  %   \label{tab:transferability}
  % \end{minipage}
  % \caption{Two tables side by side}
\end{SCtable}

For the white-box attacks, we compared to Naturalistic-Adversarial Patch \citep{hu2021naturalistic}, which uses white-box access to backpropagate the gradients through the detector and back to the GAN's latent space.
We also compared to \citet{thys2019fooling}, a white-box attack that uses both gradients and total variation loss to smooth the resultant patch. 

% \autoref{fig:digital-yolo3} shows digital attacks.
% \autoref{fig:real-yolo3_1} shows models' performance on benign inputs.
% \autoref{fig:real-yolo4_2} show qualitative results of physical attacks.
% Note that the figures show diverse situations and lighting conditions.

\begin{figure}[ht]
\centering

\begin{tabular}{cccccc}
 % \vspace*{-\baselineskip}
% \vspace{-1pt}
\textbf{BBNP} & Sparse-RS & Square & NES & RS & Latent-RS \\
% \hline
% \multicolumn{5}{c}{\textbf{Tiny-YOLOv3}} \\
\includegraphics[width=0.15\textwidth]{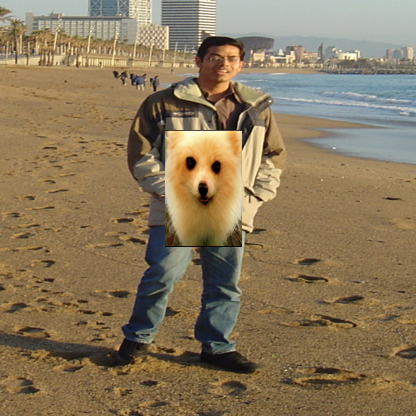} &
\includegraphics[width=0.15\textwidth]{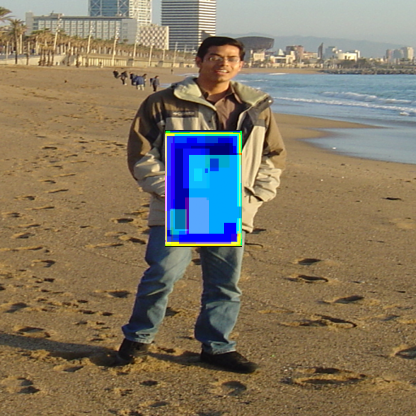} &
\includegraphics[width=0.15\textwidth]{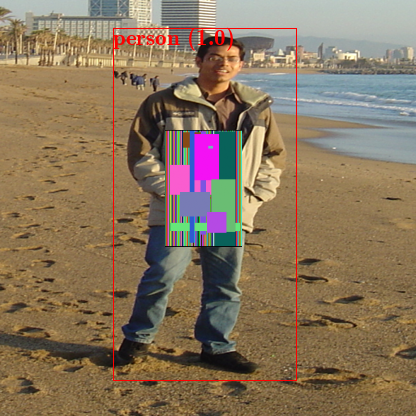} &
\includegraphics[width=0.15\textwidth]{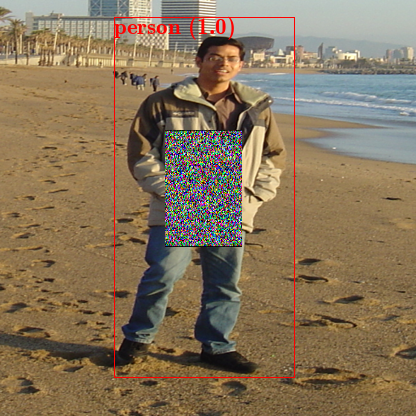} &
\includegraphics[width=0.15\textwidth]{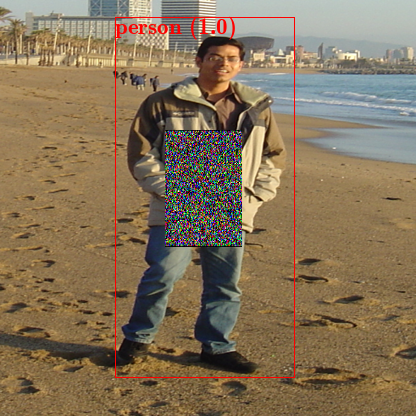} &
\includegraphics[width=0.15\textwidth]{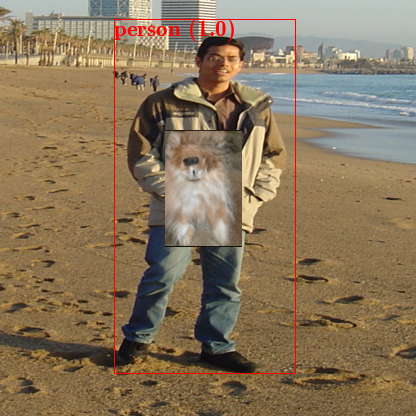}

\end{tabular}

\vspace{2pt}

\begin{tabular}{cccccc}
% \multicolumn{5}{c}{\textbf{Tiny-YOLOv4}} \\
\includegraphics[width=0.15\textwidth]{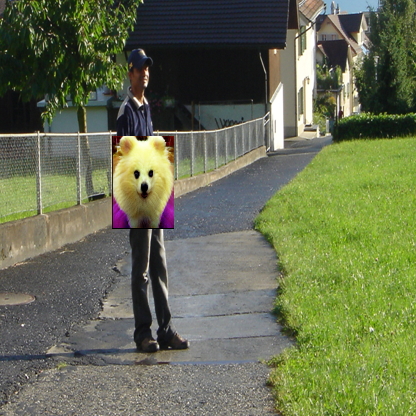} & 
\includegraphics[width=0.15\textwidth]{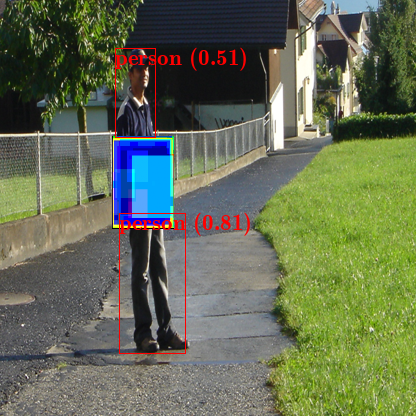} &
\includegraphics[width=0.15\textwidth]{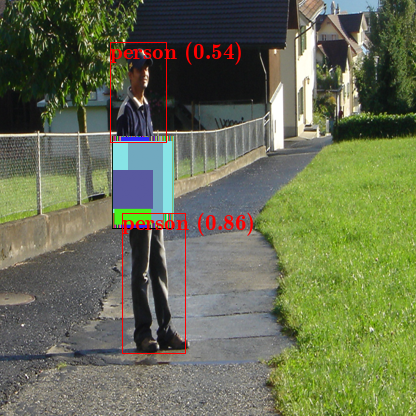} &
\includegraphics[width=0.15\textwidth]{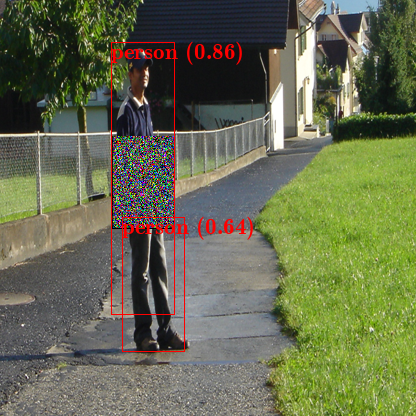} &
\includegraphics[width=0.15\textwidth]{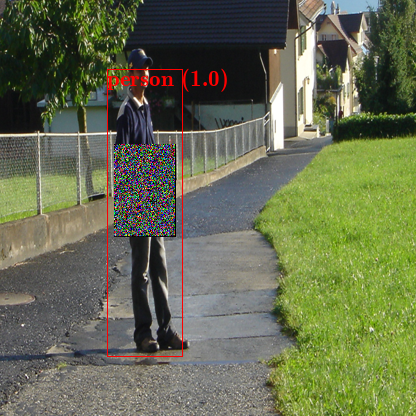} &
\includegraphics[width=0.15\textwidth]{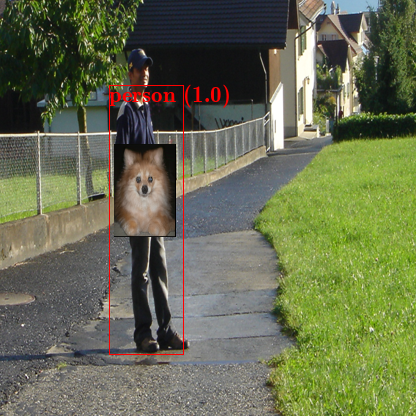}
\end{tabular}

\vspace{2pt}

\begin{tabular}{cccccc}
% \multicolumn{5}{c}{\textbf{YOLOv5s}} \\
\includegraphics[width=0.15\textwidth]{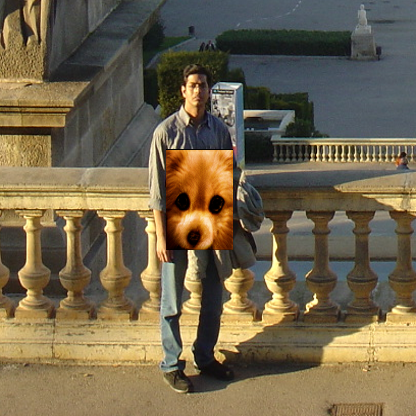} & 
\includegraphics[width=0.15\textwidth]{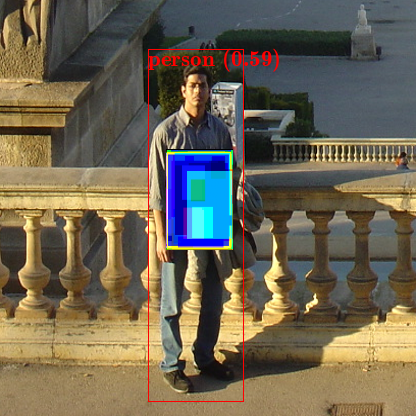} &
\includegraphics[width=0.15\textwidth]{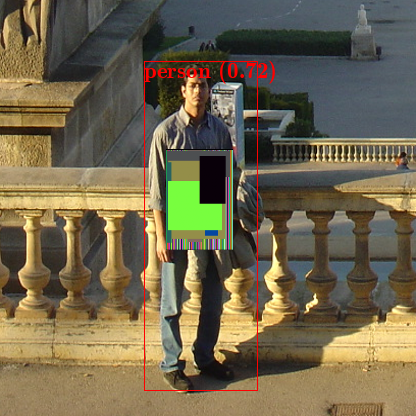} &
\includegraphics[width=0.15\textwidth]{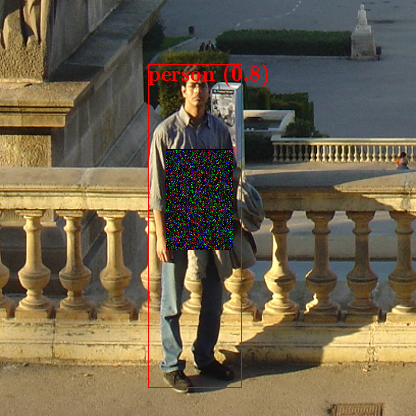} &
\includegraphics[width=0.15\textwidth]{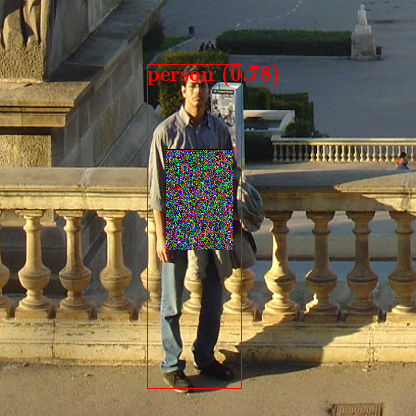} &
\includegraphics[width=0.15\textwidth]{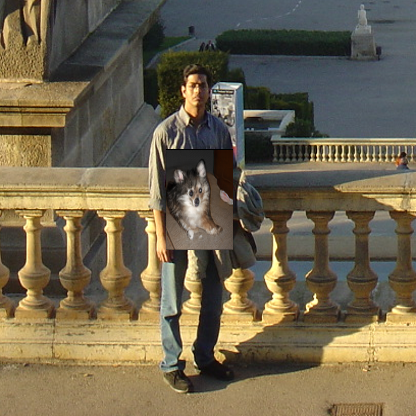} \\ 

\end{tabular}
\vspace{-10pt}
\caption{Digital examples of all tested black-box algorithms, on Tiny-YOLOv3 (top), Tiny-YOLOv4 (middle), and YOLOv5s (bottom).}
\label{fig:patch-compare}
\end{figure}

\autoref{fig:patch-compare} shows samples of our patches vs. other techniques.

\subsection{Results Analysis}
Overall, our approach outperformed \textbf{all} other black-box attacks by a wide margin (Tiny-YOLOv3: \textbf{\textcolor{ForestGreen}{-18.4}} better than 2nd-best black-box attack, 
Tiny-YOLOv4: \textbf{\textcolor{ForestGreen}{-30.8}}, 
YOLOv5s: \textbf{\textcolor{ForestGreen}{-24.9}}, SSD-L: \textbf{\textcolor{ForestGreen}{-24.7}} and L-DETR: \textbf{\textcolor{ForestGreen}{-24.9}}). The transformer-based L-DETR demonstrated the highest level of resilience against all tested attacks. Our attack was even better than the white-box approaches proposed by \citet{thys2019fooling,hu2021naturalistic} on both YOLOv5s (28.6 vs 36.1, respectively) and SSD-L (37.2 vs 43.7, respectively) and showed comparable effectiveness against Tiny-YOLOv4 (16.3 vs 14.5, respectively). Amongst the black-box attacks, Sparse-RS was the second-best attack. Not only did we demonstrate superiority against all other black-box attacks, our adversarial patches are more natural, and thus harder to detect and identify. We will elaborate on the naturalness of our attack in \autoref{sec:naturalness_assessment}. 

\subsection{Transferability Analysis}
Transferability is a crucial aspect of evaluating the robustness of black-box adversarial attacks on object detectors. Thus, we investigate the effectiveness of the proposed attack method across the different target models. 

%The experiments are designed to assess the generalization of adversarial perturbations crafted on one model, to others, thereby highlighting the potential risks and challenges associated with the transferability of adversarial attacks.

To assess transferability we picked the best generated patch on a specific model, according to \autoref{table:mAPall}, and then evaluated its effectiveness when applied to other object-detection models. The transferability of these adversarial patches was then investigated by measuring their success rates when applied to each distinct target model.
\autoref{tab:transferability} presents the transferability results across the different object-detection models. Each row corresponds to a different source model, while the columns represent the target model for the transferability analysis. The success rates of the adversarial patches are reported in terms of AP(\%). 

\begin{table}[ht]
    \centering
    \caption{Transferability results of the different adversarial patches across the different object-detection models in terms of AP (\%). Each row corresponds to a different experiment. The ``Source Model'' column shows the model that the patch was optimized on, while the other columns represent the target models. For each experiment, the best AP (lowest) reached in terms of transferability is in \textbf{bold}. The original performance values on the source model are grayed out.}
    \label{tab:transferability}
    \resizebox{0.7\linewidth}{!}{
            \renewcommand{\arraystretch}{1.1}
            \begin{tabular}{C{1.5cm} | C{2.1cm} | C{2cm} C{2cm} C{2cm} C{2cm} C{2cm}} 
            % \specialrule{.2em}{.1em}{.1em}
            Method & Source Model &
                Tiny-YOLOv3 \hspace{3px} & Tiny-YOLOv4 \hspace{3px} & YOLOv5s & SSD-L & L-DETR \\ \hline
                \textbf{BBNP} & \multirow{6}*{Tiny-YOLOv3} & \textcolor{gray!40}{23.2} & 50.9 & \textbf{36.0} & \textbf{41.6} & \textbf{77.7} \\ 
                Sparse-RS &  & \textcolor{gray!40}{41.6}  & \textbf{46.2} & 57.8 & 63.0 & 83.5 \\
                Square &  & \textcolor{gray!40}{43.9} & 66.8 & 68.5 & 64.6 & 87.7 \\
                NES &  & \textcolor{gray!40}{77.7} & 78.0 & 76.4 & 67.8 & 88.7 \\
                RS &  & \textcolor{gray!40}{79.6} & 78.1 & 76.1 & 67.3 & 88.5 \\
                Latent-RS &  & \textcolor{gray!40}{68.2}  & 67.5 & 48.1 & 65.4 & 83.7 \\
                \hline
                \textbf{BBNP} & \multirow{6}*{Tiny-YOLOv4} & \textbf{31.4} & \textcolor{gray!40}{16.3} & \textbf{36.4} & \textbf{51.9} & \textbf{80.2} \\
                Sparse-RS &  & 43.7 & \textcolor{gray!40}{47.1} & 61.7 & 63.2 & 83.0 \\
                Square &  & 59.0 & \textcolor{gray!40}{61.6} & 51.4  & 63.5 & 86.4 \\
                NES &  & 81.9 & \textcolor{gray!40}{77.9} & 76.1 & 68.0 & 89.3  \\
                RS &  & 78.5 & \textcolor{gray!40}{79.6} & 75.1  & 67.8 & 88.1 \\
                Latent-RS &  & 79.2 & \textcolor{gray!40}{67.1} & 62.8 & 67.6 & 86.4  \\
                \hline
                \textbf{BBNP} & \multirow{6}*{YOLOv5s} & 41.2 & 57.7 & \textcolor{gray!40}{28.6} & \textbf{59.3} & \textbf{77.1} \\
                Sparse-RS &  & 33.9 & \textbf{41.4} & \textcolor{gray!40}{63.0} & 62.2 & 82.7 \\
                Square &  & \textbf{32.5} & 48.4 & \textcolor{gray!40}{61.7} & 61.3 & 85.6 \\
                NES &  & 83.2 & 78.5 & \textcolor{gray!40}{71.8} & 67.5 & 89.2 \\
                RS &  & 79.4 & 78.4 & \textcolor{gray!40}{73.6} & 67.5 & 88.6 \\
                Latent-RS &  & 66.1 & 64.6 & \textcolor{gray!40}{53.5} & 60.1 & 84.3 \\

                \hline

                \textbf{BBNP} & \multirow{6}*{SSD-L} & \textbf{48.7} & 50.1 & \textbf{46.3} & \textcolor{gray!40}{37.2} & \textbf{76.7} \\
                Sparse-RS &  & 70.4 & \textbf{43.6} & 84.5 & \textcolor{gray!40}{64.0} & 82.5 \\
                Square &  & 72.0 & 57.8 & 83.5 & \textcolor{gray!40}{62.8} & 87.8 \\
                NES &  & 73.4 & 70.5 & 84.0  & \textcolor{gray!40}{67.2} & 88.5 \\
                RS &  & 72.8 & 70.3 & 85.4 & \textcolor{gray!40}{67.4} & 88.3 \\
                Latent-RS &  & 73.0 & 68.3 & 58.7 & \textcolor{gray!40}{61.9} & 81.8 \\

                \hline

                \textbf{BBNP} & \multirow{6}*{L-DETR} & \textbf{45.6} & 44.1 & \textbf{45.8} & \textbf{56.0} & \textcolor{gray!40}{57.3} \\
                Sparse-RS &  & 70.3 & \textbf{39.3} & 84.4 & 61.8 & \textcolor{gray!40}{82.2} \\
                Square &  & 70.1 & 55.8 & 84.6 & 62.4 & \textcolor{gray!40}{86.3} \\
                NES &  & 72.4 & 69.5 & 85.4 & 67.5 & \textcolor{gray!40}{87.7} \\
                RS &  & 73.1 & 69.7 & 84.6 & 67.8 & \textcolor{gray!40}{88.7} \\
                Latent-RS &  & 75.7 & 70.2 & 70.8 & 66.8 & \textcolor{gray!40}{87.0} \\

              % \specialrule{.2em}{.1em}{.1em}
                
        \end{tabular}}
        \renewcommand{\arraystretch}{\originalarraystretch} % Reset to saved value
\end{table}

Our approach demonstrated superior transferability, particularly with patches optimized on Tiny-YOLOv3 or Tiny-YOLOv4, achieving high success rates (AP\%) on YOLOv5s, indicating shared vulnerabilities across YOLO-based architectures. Notably, a patch optimized on Tiny-YOLOv4 showed the most generalizability when transferred to Tiny-YOLOv3 and YOLOv5s. Additionally, \textbf{BBNP} patches optimized on SSD-L performed well on Tiny-YOLOv3 and Tiny-YOLOv5, while patches optimized on L-DETR achieved the highest success rates on Tiny-YOLOv3, YOLOv5s and SSD-L. These findings highlight the effectiveness of our approach in generating transferable adversarial patches, potentially generalizing to unseen models with similar architectural features.

\subsection{Subjective Assessment of the Naturalness of Various Adversarial Patches}
\label{sec:naturalness_assessment}
The primary focus of our proposed methodology centers on the assessment of the naturalness and conspicuousness exhibited by generated adversarial patches, as perceived by human observers. To this end, we conducted a formalized series of subjective evaluations aimed at gauging our patch naturalness in comparison both to baseline and authentic images. The comparison involved the administration of a subjective survey to a cohort of 40 randomly chosen participants. The survey presented patches to participants in a randomized order. For benchmarking purposes, we generated 3 adversarial patches and compiled 6 more off-the-shelf adversarial patches—--three per method---generated by \citep{wu2020making,hu2021naturalistic}. 

% In the context of our digital experiments it is imperative to note that a comparative naturalness analysis of our proposed patches against other black-box methods, previously employed for benchmarking, has never been undertaken. This lacuna likely stems from the inherent non-natural characteristics shared by the other black-box methodologies. Consequently, to circumvent the risk of engendering an unfair comparison, we have refrained from juxtaposing our patches against these methodologies in the current study. 

\begin{figure}[htbp]
    \centering
    \begin{minipage}[t]{0.48\textwidth}
        \centering
        
        \caption{Naturalness subjective assessments of our adversarial patches in comparison to other baseline methodologies. The naturalness scores represent the participant votes for each test image relative to the entire cohort. As evidenced by the results, our patches garnered a higher mean score than their counterparts.}
        \begin{tabular}{c|ccc}
            % \toprule
            Source & \textbf{BBNP} & \citep{thys2019fooling} &  \citep{hu2021naturalistic} \\
            \midrule
            Patches & \raisebox{-.4\height}{\includegraphics[scale=0.25]{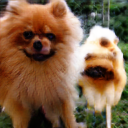}} & \raisebox{-.4\height}{\includegraphics[scale=0.25]{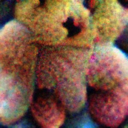}} & \raisebox{-.4\height}{\includegraphics[scale=0.25]{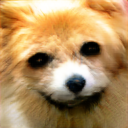}} \\
        \hline
        Naturalness & \textbf{58\%} & 30\% & 54\% \\%& 5\% \\

        % \specialrule{.1em}{.05em}{.05em}
        \midrule
        Patches & \raisebox{-.4\height}{\includegraphics[scale=0.25]{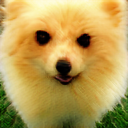}} & \raisebox{-.4\height}{\includegraphics[scale=0.25]{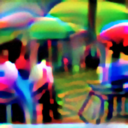}} & \raisebox{-.4\height}{\includegraphics[scale=0.25]{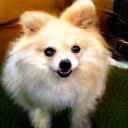}} \\
        \hline
        Naturalness & 74\% & 12\% & \textbf{100\%} \\ %& 31\%  \\

        \midrule
        Patches & \raisebox{-.4\height}{\includegraphics[scale=0.25]{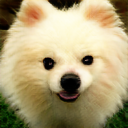}} & \raisebox{-.4\height}{\includegraphics[scale=0.25]{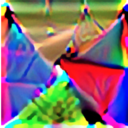}} & \raisebox{-.4\height}{\includegraphics[scale=0.25]{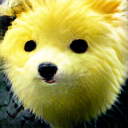}} \\ 
        \hline
        Naturalness & \textbf{85\%} & 0\% & 25\% \\ %& 9\% \\

        \hdashline[2pt/2pt]
        % \specialrule{.2em}{.1em}{.1em}
        Average & \textbf{72.3\%} & 14.6\%  & 59.6\% \\ %& 15.0\% \\
        % \specialrule{.2em}{.1em}{.1em}
            % \bottomrule
        \end{tabular}
        \label{tab:naturalness}
    \end{minipage}\hfill
    \begin{minipage}[t]{0.48\textwidth}
        \centering
        \caption{Patches evolved by our algorithm, on Tiny-YOLOv3, Tiny-YOLOv4, YOLOv5s, SSD-L and L-DETR with $\lambda_{cls}=0.2$.
Left to right: population sizes of 50, 70, 90, and 110, respectively.}
        \begin{tabular}{ccccc}
            % \toprule
        % \multicolumn{4}{c}{Tiny-YOLOv3} \\
        \parbox{1.5cm}{\centering Tiny-YOLOv3} & 
        \raisebox{-.4\height}{\includegraphics[scale=0.25]{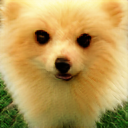}} &
        \raisebox{-.4\height}{\includegraphics[scale=0.25]{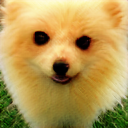}} &
        \raisebox{-.4\height}{\includegraphics[scale=0.25]{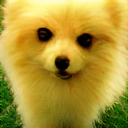}} &
        \raisebox{-.4\height}{\includegraphics[scale=0.25]{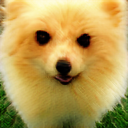}}
        
        \end{tabular}
        
        \vspace{0.55cm}
        
        \begin{tabular}{ccccc}
        % \multicolumn{4}{c}{Tiny-YOLOv4} \\
        \parbox{1.5cm}{\centering Tiny-YOLOv4} & 
        \raisebox{-.4\height}{\includegraphics[scale=0.25]{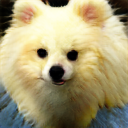}} &
        \raisebox{-.4\height}{\includegraphics[scale=0.25]{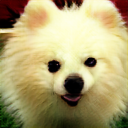}} &
        \raisebox{-.4\height}{\includegraphics[scale=0.25]{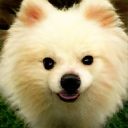}} &
        \raisebox{-.4\height}{\includegraphics[scale=0.25]{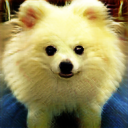}}
        \end{tabular}
        
        \vspace{0.55cm}
        
        \begin{tabular}{ccccc}
        \parbox{1.5cm}{\centering YOLOv5s} & 
        \raisebox{-.4\height}{\includegraphics[scale=0.25]{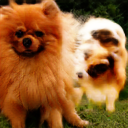}} &
        \raisebox{-.4\height}{\includegraphics[scale=0.25]{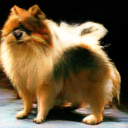}} &
        \raisebox{-.4\height}{\includegraphics[scale=0.25]{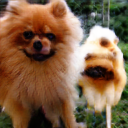}} &
        \raisebox{-.4\height}{\includegraphics[scale=0.25]{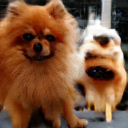}}
        \end{tabular}

        \vspace{0.55cm}
        
        \begin{tabular}{ccccc}
        \parbox{1.5cm}{\centering SSD-L} & 
        \raisebox{-.4\height}{\includegraphics[scale=0.25]{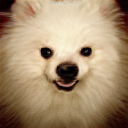}} &
        \raisebox{-.4\height}{\includegraphics[scale=0.25]{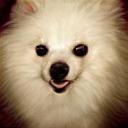}} &
        \raisebox{-.4\height}{\includegraphics[scale=0.25]{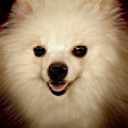}} &
        \raisebox{-.4\height}{\includegraphics[scale=0.25]{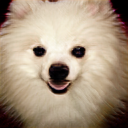}}
         \\
        \end{tabular}

        \vspace{0.55cm}

        \begin{tabular}{ccccc}
        \parbox{1.5cm}{\centering L-DETR} & 
        \raisebox{-.4\height}{\includegraphics[scale=0.25]{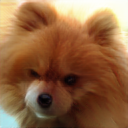}} &
        \raisebox{-.4\height}{\includegraphics[scale=0.25]{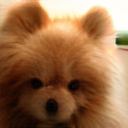}} &
        \raisebox{-.4\height}{\includegraphics[scale=0.25]{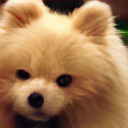}} &
        \raisebox{-.4\height}{\includegraphics[scale=0.25]{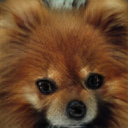}}
         \\
        \end{tabular}
        
        \label{fig:patches}
    \end{minipage}
\end{figure}

We proposed a straightforward definition of "naturalness" which is articulated as follows: "A natural image is a visual depiction of the real world, characterized by realistic scenes with genuine lighting and structure. Simply put, it refers to how authentic the image appears." Then, participants were asked to rate each patch on a scale from 1 to 7 according to its perceived naturalness, with 1 indicating the least natural and 7 indicating the most natural. We then calculated the naturalness score by determining the mean of the votes assigned to each patch and normalizing the resultant score within the range $[0,100]$. The results presented in \autoref{tab:naturalness} demonstrate that, on average, our proposed patches attain higher naturalness scores compared to the various baseline methods. \autoref{fig:patches} displays additional patches, developed for each of the targeted models.

\begin{figure}
\centering
\begin{tabular}{ccc}
\includegraphics[width = 1.7in]{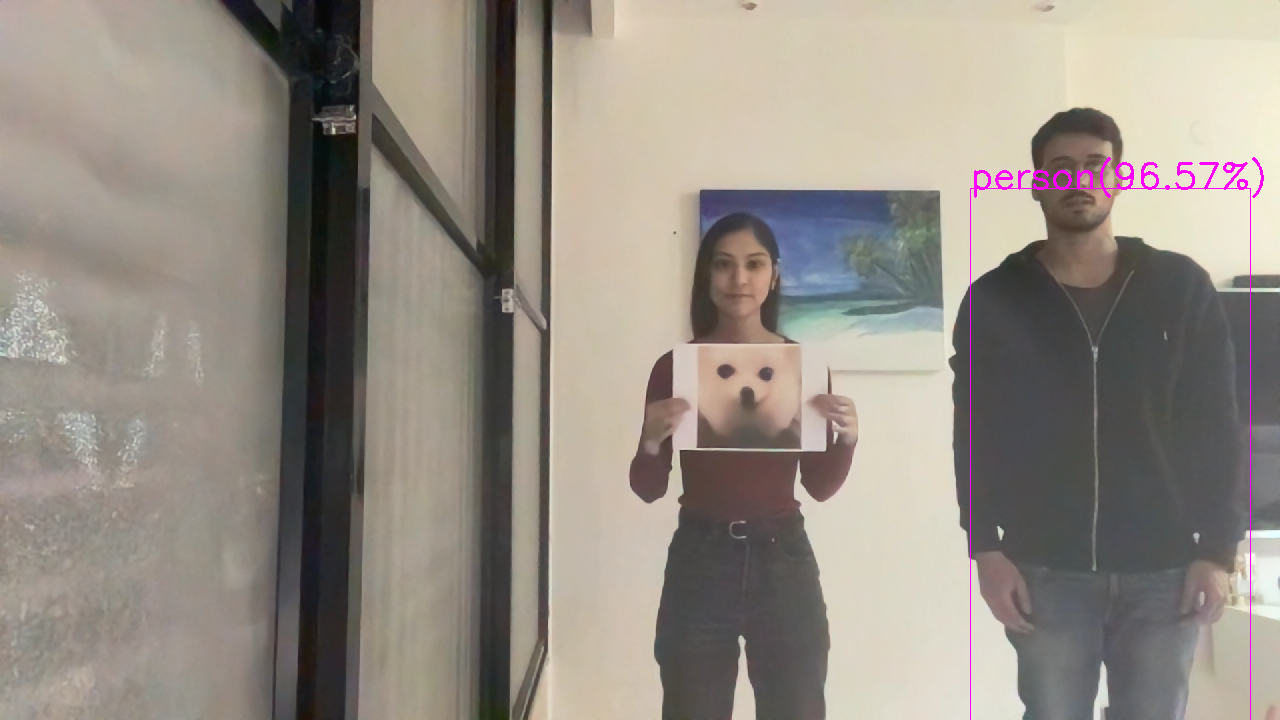} &
\includegraphics[width = 1.7in]{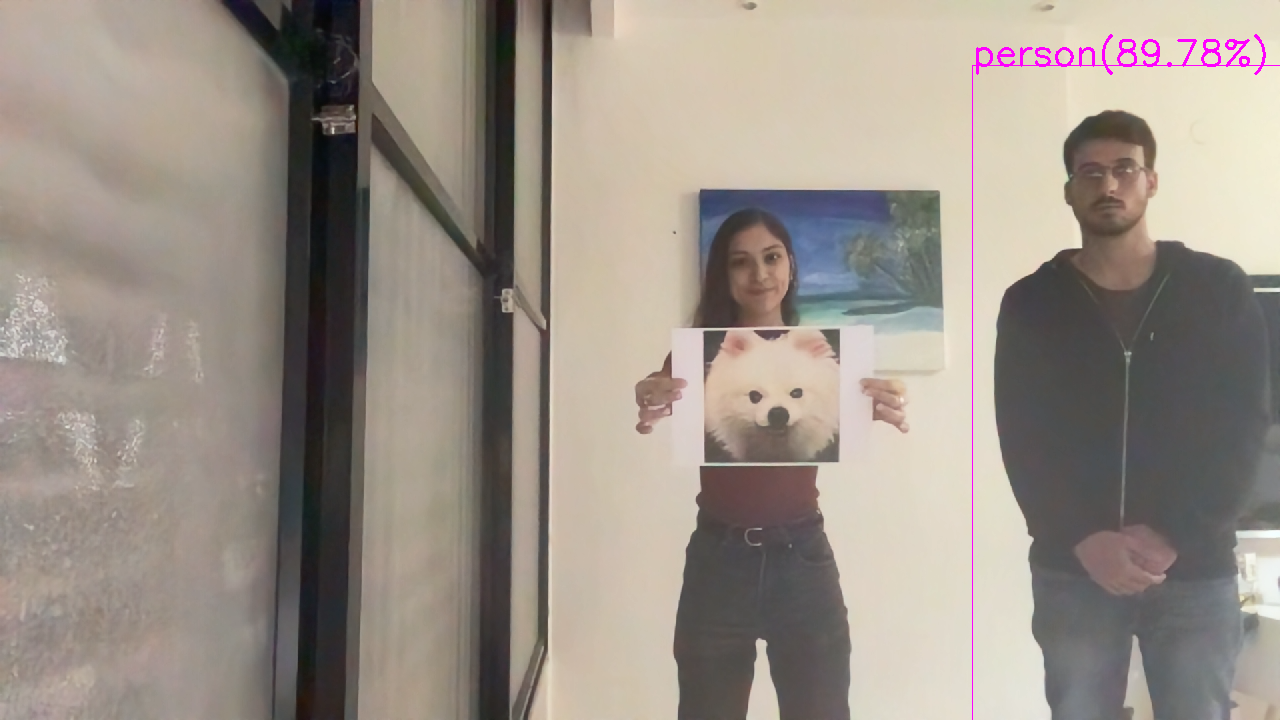} &
\includegraphics[width = 1.7in]{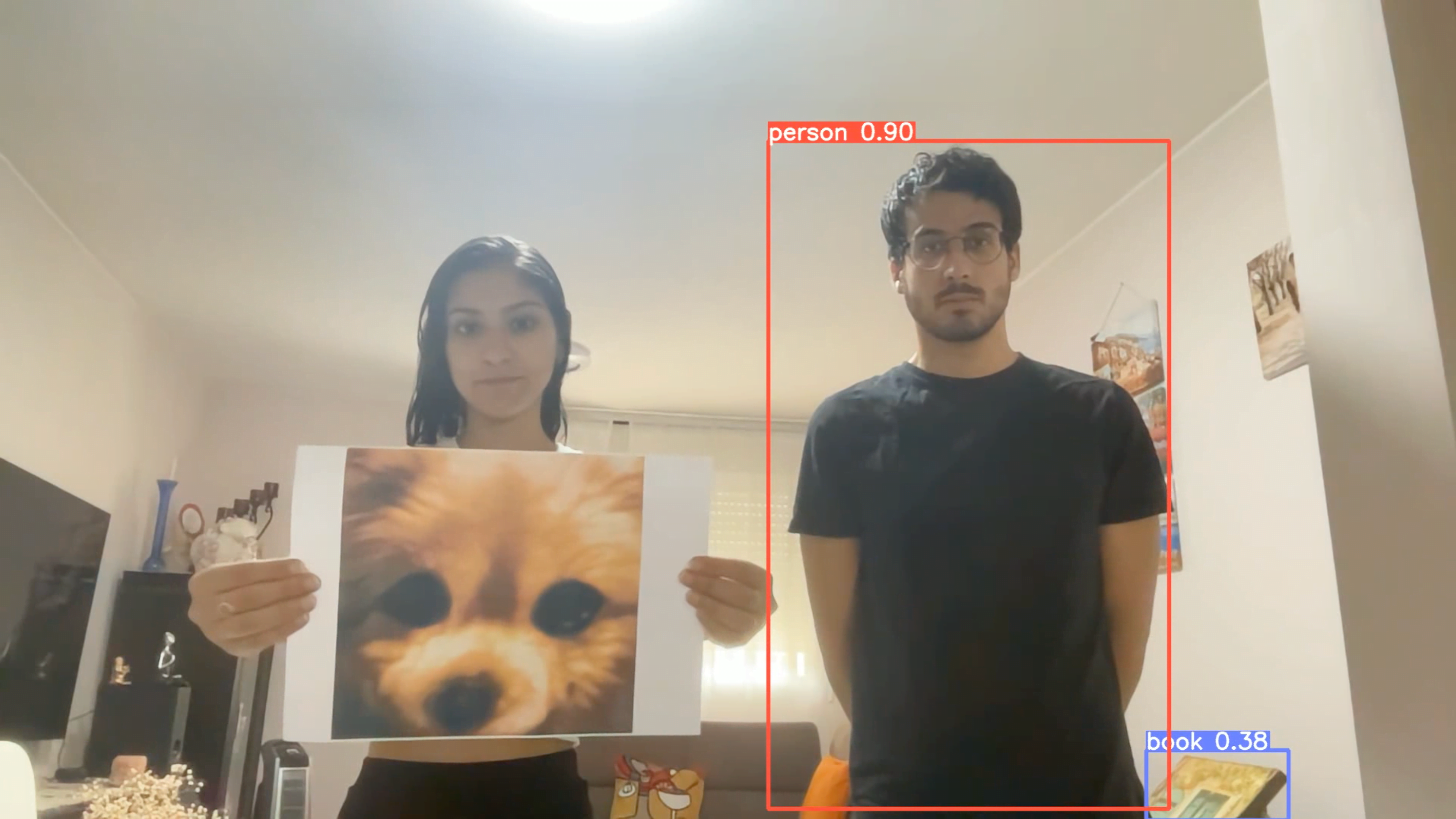}
\\
\subfloat[Tiny-YOLOv3]{\includegraphics[width = 1.7in]{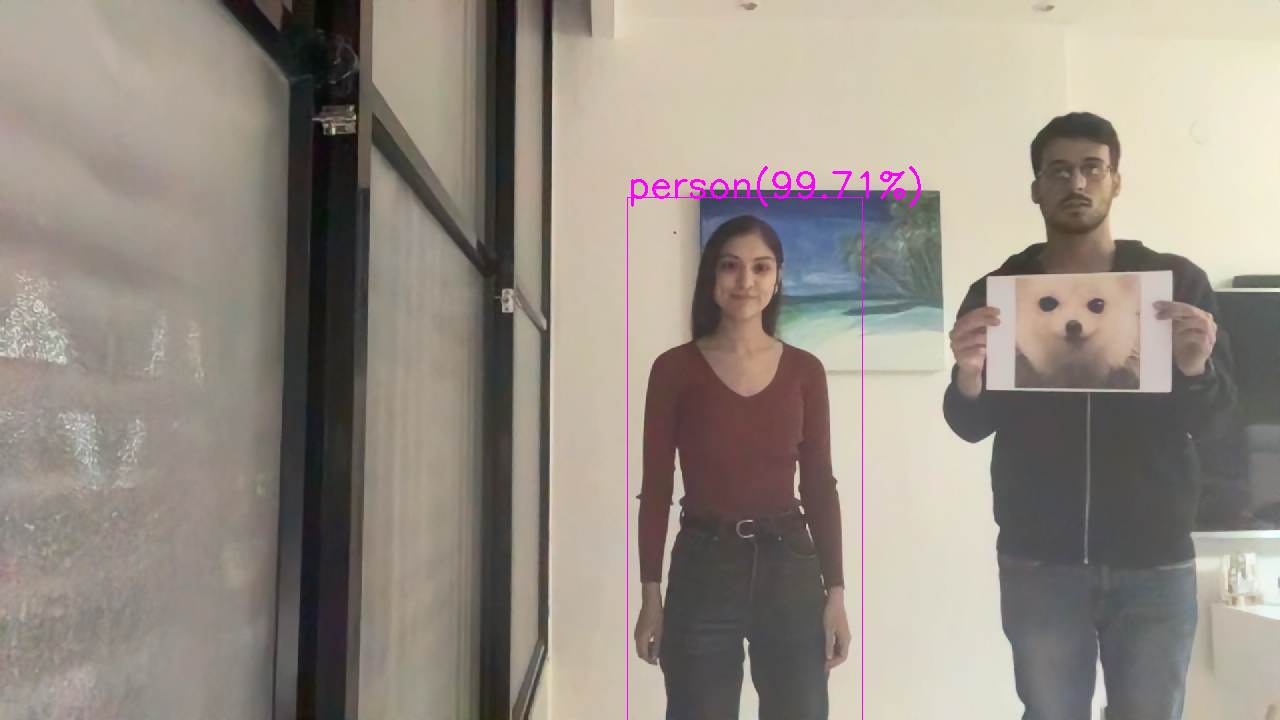}} &
\subfloat[Tiny-YOLOv4]{\includegraphics[width = 1.7in]{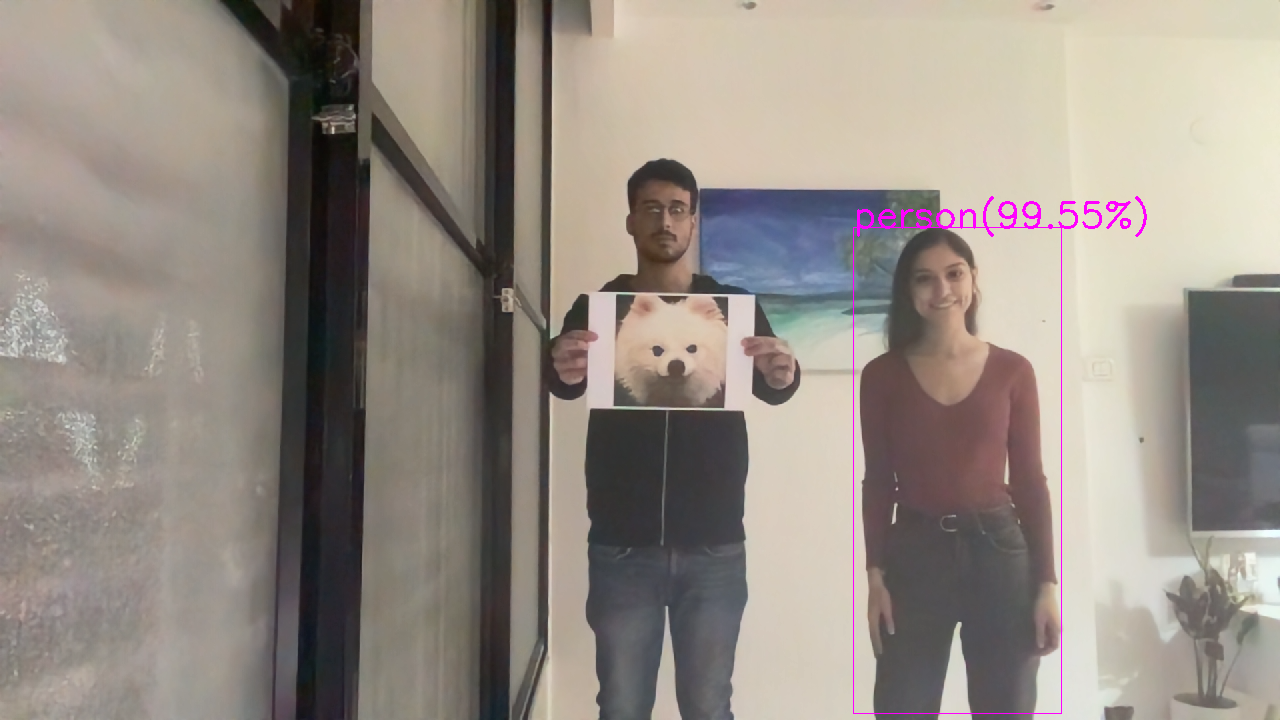}} &
\subfloat[YOLOv5s]{\includegraphics[width = 1.7in]{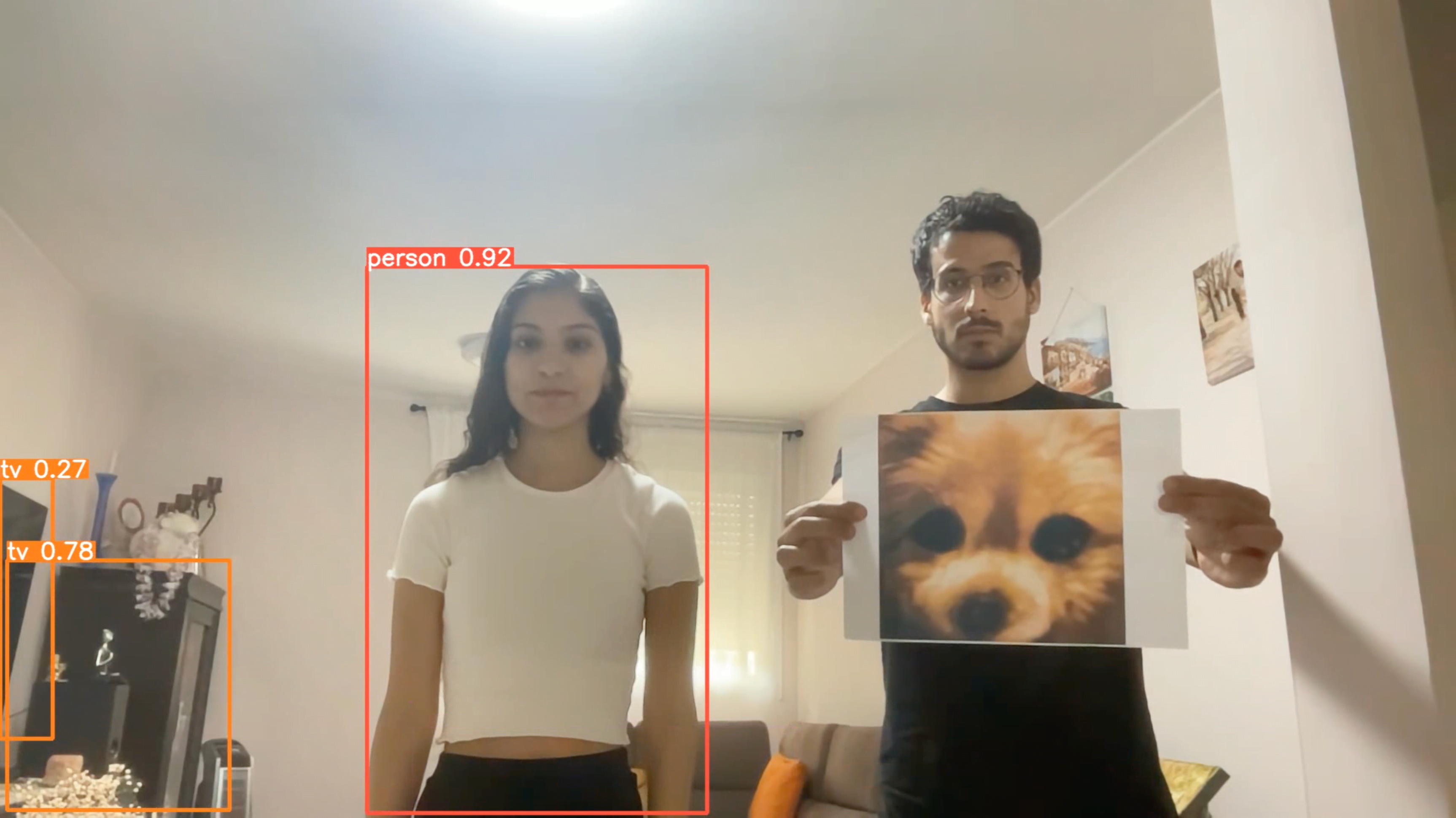}}\\
\end{tabular}
\vspace{-7pt}
\caption{Qualitative results of our evolved patches, which conceal people from object detectors.}
\label{qualitative-results}
\end{figure}

\subsection{Ablation Studies}
\label{subsec:ablation}
\begin{figure}[ht]
    \centering
    \includegraphics[scale=0.3]{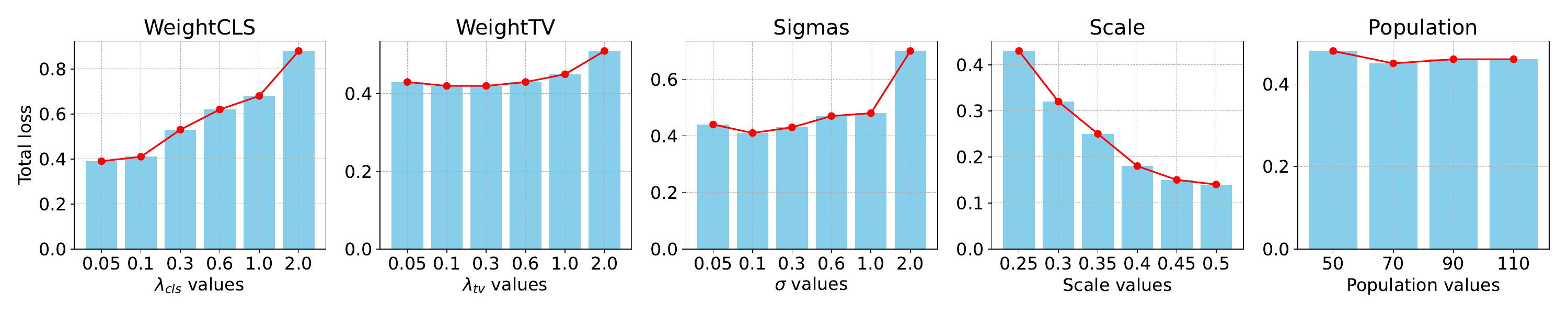}
\vspace{-7pt}
    \caption{Ablation studies using Tiny-YOLOv3 with different $\lambda_{cls}$, $\lambda_{tv}$, $\sigma$, scaling and population sizes, from left to right.}
    \label{fig:threegraphs}
\end{figure}

In this section, we perform ablation studies to analyze the impact of key hyperparameters in our loss function on the performance of our adversarial patches. Specifically, we focus on: $\lambda_{cls}$, $\lambda_{tv}$, $\sigma$, scaling sizes and population sizes. The $\lambda_{cls}$ hyperparameter scales the classification loss, which is crucial for maximizing the misclassification rate of the object detector, while the $\lambda_{tv}$ hyperparameter scales the total variation loss, which is intended to promote the smoothness and realism of the generated adversarial patches. To evaluate the effectiveness and trade-offs associated with these parameters, we conducted a series of experiments using the Tiny-YOLOv3 object detector. By systematically varying $\lambda_{cls}$ and $\lambda_{tv}$, we aim to understand their individual and combined effects on the attack success rate, visual quality of the patches, and computational efficiency. The results are depicted in \autoref{fig:threegraphs}. The $\sigma$ value regulates the exploration within the latent space---lower $\sigma$ values emphasize exploitation, while higher $\sigma$ values enhance exploration. The magnitude of the scale value dictates the proportion of the patch relative to the bounding box—lower scale values correspond to smaller patch sizes, and vice versa. It's intuitive that increasing the patch size leads to a decrease in the total loss, as illustrated in \autoref{fig:threegraphs}. We examined population sizes of 50, 70, 90, and 110. To adequately estimate gradients, we require 2 queries per coordinate of the 120-dimensional latent vector, totaling 240 queries per gradient estimation. Despite this, we utilized fewer queries, underscoring the efficacy of our approach. Our empirical findings indicate that populations of 70 and 90 yielded lower loss (\autoref{table:mAPall}). We attribute this to smaller populations enhancing algorithmic exploration, particularly beneficial in initial optimization stages, while learning rate reduction acts as an exploitation mechanism, thereby balancing optimization and resulting in reduced loss.

The ablation studies depicted in \autoref{fig:threegraphs} provide key insights into the influence of hyperparameters on the efficacy of adversarial patches in a black-box setting. The results indicate that increasing \(\lambda_{cls}\) and \(\lambda_{tv}\) enhances attack performance by placing more emphasis on classification loss and patch smoothness, respectively. Additionally, higher values of \(\sigma\) improve generalizability, while smaller scale values result in more effective patches, likely due to their compactness. Conversely, the population size shows a marginal effect, suggesting that moderate sizes may balance the search space without diminishing returns. These findings underscore the importance of careful hyperparameter tuning to optimize the robustness and efficacy of adversarial patches.

% \newpage
\section{Discussion}
\label{sec:discussion}

\subsection{Ethical Considerations and Societal Impact}
This research highlights object-detector vulnerabilities and the potential negative societal impacts they pose. Malicious use of crafted patches could compromise security systems, autonomous vehicles, and access control, raising ethical concerns such as fabricated evidence and altered advertisements, eroding trust in AI. By emphasizing these vulnerabilities, we aim to foster open communication, responsible research practices, and collaborative efforts toward developing robust and ethical AI systems. Acknowledging these potential negative impacts ensures our research contributes positively to the responsible development and deployment of object-detection technologies.

\subsection{Naturalistic Setting}
To improve the natural appearance of generated patches, several strategies can be employed. Integrating loss functions for brightness and color consistency helps the patches blend better with the background \citep{elnekave2022generating}. Additionally, texture synthesis techniques using GANs or Variational Autoencoders can create patches with realistic textures \citep{zhang2021image,xian2018texturegan}. Image processing techniques, such as Gaussian smoothing and edge-preserving filters, can refine the patches' appearance \citep{wink2004denoising}. While these methods could significantly enhance the visual integration of adversarial patches, we have focused on incorporating only the essential losses due to the challenges posed by the black-box nature of the target model. This constraint limits our ability to fully exploit these techniques in practice.

\subsection{Limitations}
\label{sec:limitations}
The study's focus on specific detector architectures limits generalizability, as patch effectiveness may vary across models and training data. Physical constraints like patch size and material may hinder real-world use. Static scenarios may not reflect dynamic conditions such as object movement or lighting changes. Countermeasures like adversarial training weren't investigated, leaving their potential efficacy uncertain. Broader ethical considerations and long-term implications for computer vision require further discussion. Acknowledging these limitations emphasizes the need for ongoing research to refine attack techniques, assess generalizability, and develop robust defenses for responsible object-detection technology deployment.

Additionally, our method assumes that the attacker has access to the classification confidence scores of the target model, which is a plausible scenario in several contexts. For instance, many machine learning APIs provide confidence scores along with predictions, enabling attackers to exploit this information for crafting adversarial examples. However, we acknowledge that this assumption might not always hold true, which limits the applicability of our approach. This is a recognized limitation, and future work could explore methods that do not rely on confidence scores, thereby broadening the scenarios where our technique can be applied.

\section{Concluding remarks}
\label{sec:concluding}
We presented a novel black-box algorithm that generates naturalistic adversarial patches for object detectors. Our approach involved optimizing a latent vector to minimize probabilities associated with the appearance of a person in the detector's output, without using any internal information of the model---a realistic scenario that does not rely on the use of gradients. We compared five different deep object detectors, five black box attacks and two white box attacks, concluding that is possible to generate patches that fool object detectors, without the use of any model internals---thus raising a serious risk. Specifically, our approach outperformed every black-box approach tested in the experiments. Empirical observations indicate that among the evaluated models, TinyYOLOv4 exhibited the lowest level of robustness, whereas L-DETR demonstrated the highest degree of robustness.  In future work we wish to explore the limitations mentioned in \autoref{sec:limitations}. 
Our forthcoming research endeavors will focus on the manifestation of these attacks within real-world settings.

\section*{Acknowledgements}
This research was partially supported by the Israeli Innovation Authority through the Trust.AI consortium.

% \clearpage
% \newpage
\small
\begingroup
\let\clearpage\relax % Prevents the bibliography from starting on a new page
\renewcommand{\bibsection}{}
\section*{Bibliography}
\bibliographystyle{unsrtnat}
\bibliography{main}

\begin{thebibliography}{54}
\providecommand{\natexlab}[1]{#1}
\providecommand{\url}[1]{\texttt{#1}}
\expandafter\ifx\csname urlstyle\endcsname\relax
  \providecommand{\doi}[1]{doi: #1}\else
  \providecommand{\doi}{doi: \begingroup \urlstyle{rm}\Url}\fi

\bibitem[Szegedy et~al.(2013)Szegedy, Zaremba, Sutskever, Bruna, Erhan,
  Goodfellow, and Fergus]{szegedy2013intriguing}
Christian Szegedy, Wojciech Zaremba, Ilya Sutskever, Joan Bruna, Dumitru Erhan,
  Ian Goodfellow, and Rob Fergus.
\newblock Intriguing properties of neural networks.
\newblock \emph{arXiv preprint arXiv:1312.6199}, 2013.

\bibitem[Carlini and Wagner(2017)]{carlini2017towards}
Nicholas Carlini and David Wagner.
\newblock Towards evaluating the robustness of neural networks.
\newblock In \emph{2017 IEEE Symposium on Security and Privacy (SP)}, pages
  39--57. Ieee, 2017.

\bibitem[Moosavi-Dezfooli et~al.(2016)Moosavi-Dezfooli, Fawzi, and
  Frossard]{moosavi2016deepfool}
Seyed-Mohsen Moosavi-Dezfooli, Alhussein Fawzi, and Pascal Frossard.
\newblock Deepfool: a simple and accurate method to fool deep neural networks.
\newblock In \emph{Proceedings of the IEEE Conference on Computer Vision and
  Pattern Recognition}, pages 2574--2582, 2016.

\bibitem[Lapid et~al.(2022)Lapid, Haramaty, and Sipper]{lapid2022evolutionary}
Raz Lapid, Zvika Haramaty, and Moshe Sipper.
\newblock An evolutionary, gradient-free, query-efficient, black-box algorithm
  for generating adversarial instances in deep convolutional neural networks.
\newblock \emph{Algorithms}, 15\penalty0 (11):\penalty0 407, 2022.

\bibitem[Lapid and Sipper(2023)]{lapid2023see}
Raz Lapid and Moshe Sipper.
\newblock I see dead people: Gray-box adversarial attack on image-to-text
  models.
\newblock In \emph{5th Workshop on Machine Learning for Cybersecurity, part of
  ECMLPKDD 2023}, 2023.

\bibitem[Thys et~al.(2019)Thys, Van~Ranst, and Goedem{\'e}]{thys2019fooling}
Simen Thys, Wiebe Van~Ranst, and Toon Goedem{\'e}.
\newblock Fooling automated surveillance cameras: Adversarial patches to attack
  person detection.
\newblock In \emph{Proceedings of the IEEE/CVF Conference on Computer Vision
  and Pattern Recognition Workshops}, pages 0--0, 2019.

\bibitem[Tamam et~al.(2023)Tamam, Lapid, and Sipper]{vitracktamam2023foiling}
Snir~Vitrack Tamam, Raz Lapid, and Moshe Sipper.
\newblock Foiling explanations in deep neural networks.
\newblock \emph{Transactions on Machine Learning Research}, 2023.
\newblock ISSN 2835-8856.
\newblock URL \url{https://openreview.net/forum?id=wvLQMHtyLk}.

\bibitem[Lapid et~al.(2024)Lapid, Langberg, and Sipper]{lapid2024open}
Raz Lapid, Ron Langberg, and Moshe Sipper.
\newblock Open sesame! universal black-box jailbreaking of large language
  models.
\newblock In \emph{ICLR 2024 Workshop on Secure and Trustworthy Large Language
  Models}, 2024.
\newblock URL \url{https://openreview.net/forum?id=0SuyNOncxX}.

\bibitem[Eykholt et~al.(2018)Eykholt, Evtimov, Fernandes, Li, Rahmati, Xiao,
  Prakash, Kohno, and Song]{eykholt2018robust}
Kevin Eykholt, Ivan Evtimov, Earlence Fernandes, Bo~Li, Amir Rahmati, Chaowei
  Xiao, Atul Prakash, Tadayoshi Kohno, and Dawn Song.
\newblock Robust physical-world attacks on deep learning visual classification.
\newblock In \emph{Proceedings of the IEEE Conference on Computer Vision and
  Pattern Recognition}, pages 1625--1634, 2018.

\bibitem[Hu et~al.(2021)Hu, Kung, Tan, Chen, Hua, and
  Cheng]{hu2021naturalistic}
Yu-Chih-Tuan Hu, Bo-Han Kung, Daniel~Stanley Tan, Jun-Cheng Chen, Kai-Lung Hua,
  and Wen-Huang Cheng.
\newblock Naturalistic physical adversarial patch for object detectors.
\newblock In \emph{Proceedings of the IEEE/CVF International Conference on
  Computer Vision}, pages 7848--7857, 2021.

\bibitem[Kurakin et~al.(2018)Kurakin, Goodfellow, and
  Bengio]{kurakin2018adversarial}
Alexey Kurakin, Ian~J Goodfellow, and Samy Bengio.
\newblock Adversarial examples in the physical world.
\newblock In \emph{Artificial Intelligence Safety and Security}, pages 99--112.
  Chapman and Hall/CRC, 2018.

\bibitem[Feng et~al.(2021)Feng, Wu, Zhang, Zhang, and Zhang]{feng2021meta}
Weiwei Feng, Baoyuan Wu, Tianzhu Zhang, Yong Zhang, and Yongdong Zhang.
\newblock Meta-attack: Class-agnostic and model-agnostic physical adversarial
  attack.
\newblock In \emph{Proceedings of the IEEE/CVF International Conference on
  Computer Vision}, pages 7787--7796, 2021.

\bibitem[Zolfi et~al.(2021)Zolfi, Avidan, Elovici, and
  Shabtai]{zolfi2021adversarial}
Alon Zolfi, Shai Avidan, Yuval Elovici, and Asaf Shabtai.
\newblock Adversarial mask: Real-world adversarial attack against face
  recognition models.
\newblock \emph{arXiv preprint arXiv:2111.10759}, 2021.

\bibitem[Sharif et~al.(2019)Sharif, Bhagavatula, Bauer, and
  Reiter]{sharif2019general}
Mahmood Sharif, Sruti Bhagavatula, Lujo Bauer, and Michael~K Reiter.
\newblock A general framework for adversarial examples with objectives.
\newblock \emph{ACM Transactions on Privacy and Security (TOPS)}, 22\penalty0
  (3):\penalty0 1--30, 2019.

\bibitem[Redmon et~al.(2016)Redmon, Divvala, Girshick, and
  Farhadi]{redmon2016you}
Joseph Redmon, Santosh Divvala, Ross Girshick, and Ali Farhadi.
\newblock You only look once: Unified, real-time object detection.
\newblock In \emph{Proceedings of the IEEE Conference on Computer Vision and
  Pattern Recognition}, pages 779--788, 2016.

\bibitem[Liu et~al.(2018)Liu, Yang, Liu, Song, Li, and Chen]{liu2018dpatch}
Xin Liu, Huanrui Yang, Ziwei Liu, Linghao Song, Hai Li, and Yiran Chen.
\newblock Dpatch: An adversarial patch attack on object detectors.
\newblock \emph{arXiv preprint arXiv:1806.02299}, 2018.

\bibitem[Im~Choi and Tian(2022)]{im2022adversarial}
Jung Im~Choi and Qing Tian.
\newblock Adversarial attack and defense of {YOLO} detectors in autonomous
  driving scenarios.
\newblock In \emph{2022 IEEE Intelligent Vehicles Symposium (IV)}, pages
  1011--1017. IEEE, 2022.

\bibitem[Liu et~al.(2016)Liu, Chen, Liu, and Song]{liu2016delving}
Yanpei Liu, Xinyun Chen, Chang Liu, and Dawn Song.
\newblock Delving into transferable adversarial examples and black-box attacks.
\newblock \emph{arXiv preprint arXiv:1611.02770}, 2016.

\bibitem[Li et~al.(2020)Li, Deng, Li, Yan, Gao, and Huang]{li2020towards}
Maosen Li, Cheng Deng, Tengjiao Li, Junchi Yan, Xinbo Gao, and Heng Huang.
\newblock Towards transferable targeted attack.
\newblock In \emph{Proceedings of the IEEE/CVF Conference on Computer Vision
  and Pattern Recognition}, pages 641--649, 2020.

\bibitem[Zhang et~al.(2022)Zhang, Benz, Karjauv, Cho, Zhang, and
  Kweon]{zhang2022investigating}
Chaoning Zhang, Philipp Benz, Adil Karjauv, Jae~Won Cho, Kang Zhang, and In~So
  Kweon.
\newblock Investigating top-k white-box and transferable black-box attack.
\newblock In \emph{Proceedings of the IEEE/CVF Conference on Computer Vision
  and Pattern Recognition}, pages 15085--15094, 2022.

\bibitem[Qin et~al.(2021)Qin, Xiong, Yi, and Hsieh]{qin2021training}
Yunxiao Qin, Yuanhao Xiong, Jinfeng Yi, and Cho-Jui Hsieh.
\newblock Training meta-surrogate model for transferable adversarial attack.
\newblock \emph{arXiv preprint arXiv:2109.01983}, 2021.

\bibitem[Goodfellow et~al.(2020)Goodfellow, Pouget-Abadie, Mirza, Xu,
  Warde-Farley, Ozair, Courville, and Bengio]{goodfellow2020generative}
Ian Goodfellow, Jean Pouget-Abadie, Mehdi Mirza, Bing Xu, David Warde-Farley,
  Sherjil Ozair, Aaron Courville, and Yoshua Bengio.
\newblock Generative adversarial networks.
\newblock \emph{Communications of the ACM}, 63\penalty0 (11):\penalty0
  139--144, 2020.

\bibitem[Ho et~al.(2020)Ho, Jain, and Abbeel]{ho2020denoising}
Jonathan Ho, Ajay Jain, and Pieter Abbeel.
\newblock Denoising diffusion probabilistic models.
\newblock \emph{Advances in Neural Information Processing Systems},
  33:\penalty0 6840--6851, 2020.

\bibitem[Karras et~al.(2019)Karras, Laine, and Aila]{karras2019style}
Tero Karras, Samuli Laine, and Timo Aila.
\newblock A style-based generator architecture for generative adversarial
  networks.
\newblock In \emph{Proceedings of the IEEE/CVF Conference on Computer Vision
  and Pattern Recognition}, pages 4401--4410, 2019.

\bibitem[Arjovsky et~al.(2017)Arjovsky, Chintala, and
  Bottou]{arjovsky2017wasserstein}
Martin Arjovsky, Soumith Chintala, and L{\'e}on Bottou.
\newblock Wasserstein generative adversarial networks.
\newblock In \emph{International Conference on Machine Learning}, pages
  214--223. PMLR, 2017.

\bibitem[Mao et~al.(2017)Mao, Li, Xie, Lau, Wang, and
  Paul~Smolley]{mao2017least}
Xudong Mao, Qing Li, Haoran Xie, Raymond~YK Lau, Zhen Wang, and Stephen
  Paul~Smolley.
\newblock Least squares generative adversarial networks.
\newblock In \emph{Proceedings of the IEEE International Conference on Computer
  Vision}, pages 2794--2802, 2017.

\bibitem[Brock et~al.(2018)Brock, Donahue, and Simonyan]{brock2018large}
Andrew Brock, Jeff Donahue, and Karen Simonyan.
\newblock Large scale gan training for high fidelity natural image synthesis.
\newblock \emph{arXiv preprint arXiv:1809.11096}, 2018.

\bibitem[Brown et~al.(2017)Brown, Man{\'e}, Roy, Abadi, and
  Gilmer]{brown2017adversarial}
Tom~B Brown, Dandelion Man{\'e}, Aurko Roy, Mart{\'\i}n Abadi, and Justin
  Gilmer.
\newblock Adversarial patch.
\newblock \emph{arXiv preprint arXiv:1712.09665}, 2017.

\bibitem[Lee and Kolter(2019)]{lee2019physical}
Mark Lee and Zico Kolter.
\newblock On physical adversarial patches for object detection.
\newblock \emph{arXiv preprint arXiv:1906.11897}, 2019.

\bibitem[Wierstra et~al.(2014)Wierstra, Schaul, Glasmachers, Sun, Peters, and
  Schmidhuber]{wierstra2014natural}
Daan Wierstra, Tom Schaul, Tobias Glasmachers, Yi~Sun, Jan Peters, and
  J{\"u}rgen Schmidhuber.
\newblock Natural evolution strategies.
\newblock \emph{Journal of Machine Learning Research}, 15\penalty0
  (1):\penalty0 949--980, 2014.

\bibitem[Deng et~al.(2009)Deng, Dong, Socher, Li, Li, and
  Fei-Fei]{deng2009imagenet}
Jia Deng, Wei Dong, Richard Socher, Li-Jia Li, Kai Li, and Li~Fei-Fei.
\newblock Imagenet: A large-scale hierarchical image database.
\newblock In \emph{2009 IEEE Conference on Computer Vision and Pattern
  Recognition}, pages 248--255. Ieee, 2009.

\bibitem[He et~al.(2016)He, Zhang, Ren, and Sun]{he2016deep}
Kaiming He, Xiangyu Zhang, Shaoqing Ren, and Jian Sun.
\newblock Deep residual learning for image recognition.
\newblock In \emph{Proceedings of the IEEE conference on computer vision and
  pattern recognition}, pages 770--778, 2016.

\bibitem[Dalal and Triggs(2005)]{dalal2005histograms}
Navneet Dalal and Bill Triggs.
\newblock Histograms of oriented gradients for human detection.
\newblock In \emph{2005 IEEE Computer Society Conference on Computer Vision and
  Pattern Recognition (CVPR'05)}, volume~1, pages 886--893. Ieee, 2005.

\bibitem[Croce et~al.(2022)Croce, Andriushchenko, Singh, Flammarion, and
  Hein]{croce2022sparse}
Francesco Croce, Maksym Andriushchenko, Naman~D Singh, Nicolas Flammarion, and
  Matthias Hein.
\newblock Sparse-rs: a versatile framework for query-efficient sparse black-box
  adversarial attacks.
\newblock In \emph{Proceedings of the AAAI Conference on Artificial
  Intelligence}, volume~36, pages 6437--6445, 2022.

\bibitem[Andriushchenko et~al.(2020)Andriushchenko, Croce, Flammarion, and
  Hein]{andriushchenko2020square}
Maksym Andriushchenko, Francesco Croce, Nicolas Flammarion, and Matthias Hein.
\newblock Square attack: a query-efficient black-box adversarial attack via
  random search.
\newblock In \emph{European conference on computer vision}, pages 484--501.
  Springer, 2020.

\bibitem[Ilyas et~al.(2018)Ilyas, Engstrom, Athalye, and Lin]{ilyas2018black}
Andrew Ilyas, Logan Engstrom, Anish Athalye, and Jessy Lin.
\newblock Black-box adversarial attacks with limited queries and information.
\newblock In \emph{International conference on machine learning}, pages
  2137--2146. PMLR, 2018.

\bibitem[Bergstra and Bengio(2012)]{bergstra2012random}
James Bergstra and Yoshua Bengio.
\newblock Random search for hyper-parameter optimization.
\newblock \emph{Journal of machine learning research}, 13\penalty0 (2), 2012.

\bibitem[Kingma and Ba(2014)]{kingma2014adam}
Diederik~P Kingma and Jimmy Ba.
\newblock Adam: A method for stochastic optimization.
\newblock \emph{arXiv preprint arXiv:1412.6980}, 2014.

\bibitem[Redmon and Farhadi(2018)]{redmon2018yolov3}
Joseph Redmon and Ali Farhadi.
\newblock Yolov3: An incremental improvement.
\newblock \emph{arXiv preprint arXiv:1804.02767}, 2018.

\bibitem[Bochkovskiy et~al.(2020)Bochkovskiy, Wang, and
  Liao]{bochkovskiy2020yolov4}
Alexey Bochkovskiy, Chien-Yao Wang, and Hong-Yuan~Mark Liao.
\newblock Yolov4: Optimal speed and accuracy of object detection.
\newblock \emph{arXiv preprint arXiv:2004.10934}, 2020.

\bibitem[Jocher(2020)]{Jocher_YOLOv5_by_Ultralytics_2020}
Glenn Jocher.
\newblock {YOLOv5 by Ultralytics}, May 2020.
\newblock URL \url{https://github.com/ultralytics/yolov5}.

\bibitem[Terven and Cordova-Esparza(2023)]{terven2304comprehensive}
J~Terven and D~Cordova-Esparza.
\newblock A comprehensive review of {YOLO}: From {YOLOv1} and beyond.
\newblock \emph{arXiv preprint arXiv:2304.00501}, 2023.

\bibitem[Howard et~al.(2019)Howard, Sandler, Chu, Chen, Chen, Tan, Wang, Zhu,
  Pang, Vasudevan, et~al.]{howard2019searching}
Andrew Howard, Mark Sandler, Grace Chu, Liang-Chieh Chen, Bo~Chen, Mingxing
  Tan, Weijun Wang, Yukun Zhu, Ruoming Pang, Vijay Vasudevan, et~al.
\newblock Searching for mobilenetv3.
\newblock In \emph{Proceedings of the IEEE/CVF International Conference on
  Computer Vision}, pages 1314--1324, 2019.

\bibitem[Sandler et~al.(2018)Sandler, Howard, Zhu, Zhmoginov, and
  Chen]{sandler2018mobilenetv2}
Mark Sandler, Andrew Howard, Menglong Zhu, Andrey Zhmoginov, and Liang-Chieh
  Chen.
\newblock Mobilenetv2: Inverted residuals and linear bottlenecks.
\newblock In \emph{Proceedings of the IEEE conference on computer vision and
  pattern recognition}, pages 4510--4520, 2018.

\bibitem[Li et~al.(2023)Li, Zeng, Liu, Zhang, Li, Zhang, and Ni]{li2023lite}
Feng Li, Ailing Zeng, Shilong Liu, Hao Zhang, Hongyang Li, Lei Zhang, and
  Lionel~M Ni.
\newblock Lite detr: An interleaved multi-scale encoder for efficient detr.
\newblock In \emph{Proceedings of the IEEE/CVF Conference on Computer Vision
  and Pattern Recognition}, pages 18558--18567, 2023.

\bibitem[Lin et~al.(2014)Lin, Maire, Belongie, Hays, Perona, Ramanan,
  Doll{\'a}r, and Zitnick]{lin2014microsoft}
Tsung-Yi Lin, Michael Maire, Serge Belongie, James Hays, Pietro Perona, Deva
  Ramanan, Piotr Doll{\'a}r, and C~Lawrence Zitnick.
\newblock Microsoft coco: Common objects in context.
\newblock In \emph{Computer Vision--ECCV 2014: 13th European Conference,
  Zurich, Switzerland, September 6-12, 2014, Proceedings, Part V 13}, pages
  740--755. Springer, 2014.

\bibitem[Feng et~al.(2022)Feng, Mu, Zhong, Zhang, and Yuan]{feng2022benchmark}
Haogang Feng, Gaoze Mu, Shida Zhong, Peichang Zhang, and Tao Yuan.
\newblock Benchmark analysis of yolo performance on edge intelligence devices.
\newblock \emph{Cryptography}, 6\penalty0 (2):\penalty0 16, 2022.

\bibitem[Ali-Gombe et~al.(2021)Ali-Gombe, Elyan, Moreno-Garc{\'\i}a, and
  Zwiegelaar]{ali2021face}
Adamu Ali-Gombe, Eyad Elyan, Carlos~Francisco Moreno-Garc{\'\i}a, and Johan
  Zwiegelaar.
\newblock Face detection with {YOLO} on edge.
\newblock In \emph{International Conference on Engineering Applications of
  Neural Networks}, pages 284--292. Springer, 2021.

\bibitem[Gupta et~al.(2022)Gupta, Pareek, Singal, and Rao]{gupta2022edge}
Priyanka Gupta, Bhavya Pareek, Gaurav Singal, and D~Vijay Rao.
\newblock Edge device based military vehicle detection and classification from
  {UAV}.
\newblock \emph{Multimedia Tools and Applications}, pages 1--22, 2022.

\bibitem[Wu et~al.(2020)Wu, Lim, Davis, and Goldstein]{wu2020making}
Zuxuan Wu, Ser-Nam Lim, Larry~S Davis, and Tom Goldstein.
\newblock Making an invisibility cloak: Real world adversarial attacks on
  object detectors.
\newblock In \emph{Computer Vision--ECCV 2020: 16th European Conference,
  Glasgow, UK, August 23--28, 2020, Proceedings, Part IV 16}, pages 1--17.
  Springer, 2020.

\bibitem[Elnekave and Weiss(2022)]{elnekave2022generating}
Ariel Elnekave and Yair Weiss.
\newblock Generating natural images with direct patch distributions matching.
\newblock In \emph{European Conference on Computer Vision}, pages 544--560.
  Springer, 2022.

\bibitem[Zhang and Liu(2021)]{zhang2021image}
Libao Zhang and Yanan Liu.
\newblock Image generation based on texture guided vae-agan for regions of
  interest detection in remote sensing images.
\newblock In \emph{ICASSP 2021-2021 IEEE International Conference on Acoustics,
  Speech and Signal Processing (ICASSP)}, pages 2310--2314. IEEE, 2021.

\bibitem[Xian et~al.(2018)Xian, Sangkloy, Agrawal, Raj, Lu, Fang, Yu, and
  Hays]{xian2018texturegan}
Wenqi Xian, Patsorn Sangkloy, Varun Agrawal, Amit Raj, Jingwan Lu, Chen Fang,
  Fisher Yu, and James Hays.
\newblock Texturegan: Controlling deep image synthesis with texture patches.
\newblock In \emph{Proceedings of the IEEE conference on computer vision and
  pattern recognition}, pages 8456--8465, 2018.

\bibitem[Wink and Roerdink(2004)]{wink2004denoising}
Alle~Meije Wink and Jos~BTM Roerdink.
\newblock Denoising functional mr images: a comparison of wavelet denoising and
  gaussian smoothing.
\newblock \emph{IEEE transactions on medical imaging}, 23\penalty0
  (3):\penalty0 374--387, 2004.

\end{thebibliography}
\endgroup

\end{document}